\pdfoutput=1
\documentclass[10pt,twocolumn,letterpaper]{article}

\usepackage{iccv}
\usepackage{times}
\usepackage{epsfig}
\usepackage{graphicx}
\usepackage{amsmath}
\usepackage{amssymb}

\usepackage{subfig}
\usepackage{booktabs}
\usepackage{array}
\usepackage{rotating}
\usepackage{makecell}
\usepackage{xcolor}

\usepackage{titling}
\date{}\predate{}\postdate{}
\pretitle{\begin{center}\Large \bf}
\posttitle{\end{center}\vspace{-1.8em}}

\usepackage[loop,autoplay]{animate}
\newif\ifcompileanimated
\compileanimatedtrue
\newif\ifcompilewithsupplementary

\newif\ifcompileanimatedexperimentpage
\compileanimatedexperimentpagetrue

\usepackage{adjustbox}
\usepackage{enumitem}

\usepackage[pagebackref=true,breaklinks=true,colorlinks,bookmarks=false]{hyperref}
\hypersetup{
  pdftitle={Event-Based Motion Segmentation by Motion Compensation},
  pdfsubject={Computer Vision, Robotics, Optimization},
  pdfauthor={Timo Stoffregen, Guillermo Gallego, Tom Drummond, Lindsay Kleeman, Davide Scaramuzza},
  pdfkeywords={Event cameras, Motion Segmentation, Asynchronous sensor, Motion Estimation, High Dynamic Range, High temporal resolution, Low latency}
}
\iccvfinalcopy

\def\iccvPaperID{1829}

\ificcvfinal\fi

\usepackage{cite}
\usepackage[per-mode=symbol]{siunitx}
\DeclareSIUnit\revpermin{rpm}

\usepackage{amsfonts} 

\global\long\def\cE{\mathcal{E}} 
\global\long\def\btheta{\boldsymbol{\theta}} 
\global\long\def\bparams{\btheta} 

\global\long\def\Warp{\mathbf{W}} 
\global\long\def\tref{t_\text{ref}} 
\global\long\def\pol{s} 
\global\long\def\IWE{I} 

\global\long\def\R{\mathbb{R}} 
\global\long\def\bx{\mathbf{x}}

\global\long\def\numEvents{N_e} 
\global\long\def\numPixels{N_p} 
\global\long\def\numIters{N_\text{it}} 
\global\long\def\layer{\ell} 
\global\long\def\numLayers{N_\layer} 
\global\long\def\probmat{\mathbf{P}} 

\global\long\def\bpi{\boldsymbol{\pi}} 

\global\long\def\variance{\operatorname{Var}}

\usepackage{amsmath}
\DeclareMathOperator*{\argmax}{arg\,max}

\usepackage{algorithm}
\usepackage[noend]{algpseudocode}

\def\MYTITLE{Event-Based Motion Segmentation by Motion Compensation}
\title{\MYTITLE}

\author{Timo Stoffregen$^{1,2}$, Guillermo Gallego$^3$, Tom Drummond$^{1,2}$, Lindsay Kleeman$^1$, Davide Scaramuzza$^3$\\[1ex]
$^1$Dept. Electrical and Computer Systems Engineering, Monash University, Australia.\\
$^2$Australian Centre of Excellence for Robotic Vision, Australia.\\
$^3$Dept. Informatics (Univ. Zurich) and Dept. Neuroinformatics (Univ. Zurich \& ETH Zurich), Switzerland.
}

\usepackage[normalem]{ulem}

\usepackage{balance}
\begin{document}

\twocolumn[{%
\renewcommand\twocolumn[1][]{#1}%
\centering{\textcolor{gray}{\large{This paper has been accepted for publication at the\\ IEEE International Conference on Computer Vision (ICCV), Seoul, 2019. \textcopyright IEEE }}}\vspace{-7ex}
\maketitle
\vspace{-4ex}
\begin{center}
\centering
\includegraphics[width=\linewidth]{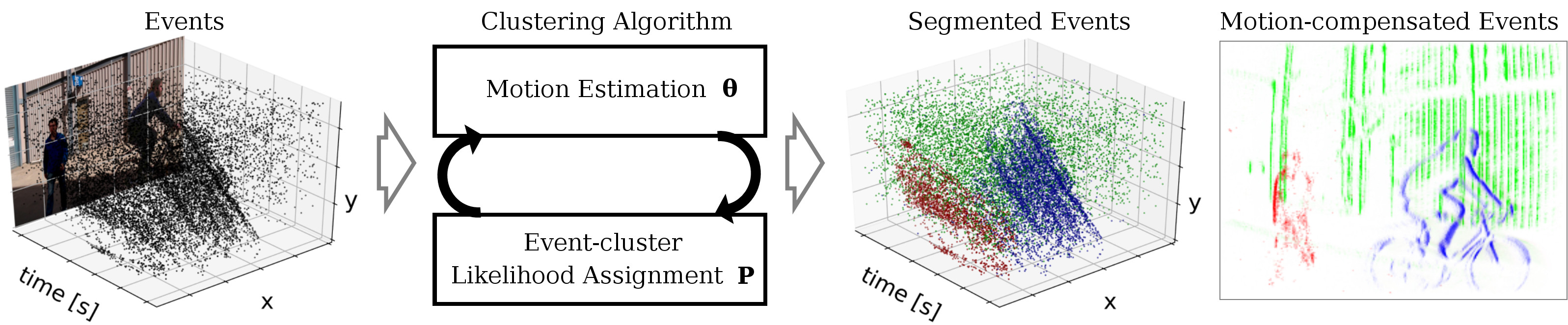}
  \captionof{figure}{\label{fig:method:diagram}%
Our method segments a set of events produced by an event-based camera (Left, with color image of the scene for illustration) 
into the different moving objects causing them (Right: pedestrian, cyclist and camera's ego-motion, in color).
We propose an iterative clustering algorithm (Middle block) that jointly estimates the motion parameters $\bparams$ and event-cluster membership probabilities $\probmat$ to best explain the scene, yielding motion-compensated event images on all clusters (Right).
}
\vspace{2ex}
\end{center}%
}]
\ificcvfinal\thispagestyle{empty}\fi

\global\long\def\widthFourSnapshots{0.15\linewidth}
\begin{abstract}
\vspace{-3ex}
In contrast to traditional cameras, whose pixels have a common exposure time, 
event-based cameras are novel bio-inspired sensors whose pixels work independently and \emph{asynchronously} output intensity changes (called ``events''), with microsecond resolution.
Since events are caused by the apparent motion of objects, event-based cameras sample visual information based on the scene dynamics and are, therefore, a more natural fit than traditional cameras to acquire motion, 
especially at high speeds, where traditional cameras suffer from motion blur.
However, distinguishing between events caused by different moving objects and by the camera's ego-motion is a challenging task.
We present the first per-event segmentation method for splitting a scene into independently moving objects.
Our method jointly estimates the event-object associations (i.e., segmentation) and the motion parameters of the objects (or the background) by maximization of an objective function, which builds upon recent results on event-based motion-compensation.
We provide a thorough evaluation of our method on a public dataset, outperforming the state-of-the-art by as much as~\SI{10}{\percent}.
We also show the first quantitative evaluation of a segmentation algorithm for event cameras, yielding around
\SI{90}{\percent} accuracy at \SI{4}{pixels} relative displacement.
\end{abstract}
\vspace{-2ex}
\section*{Supplementary Material}
\vspace{-1ex}
Accompanying video: \url{https://youtu.be/0q6ap_OSBAk}. 
We encourage the reader to view the added experiments and theory in the supplement.
\vspace{-1ex}
\section{Introduction}
\label{sec:introduction}
\vspace{-1ex}
Event-based cameras, such as the Dynamic Vision Sensor (DVS)~\cite{Lichtsteiner08ssc,Gallego19arxiv}, are novel, bio-inspired visual sensors. 
In contrast to conventional cameras that produce images at a fixed rate, the pixels in an event-based camera operate independently and asynchronously, responding to intensity changes by producing \emph{events}.
Events are represented by the $x,y$ pixel location and timestamp $t$ (in microseconds) of an intensity change as well as its polarity (i.e., whether the pixel became darker or brighter). 
Since event-based cameras essentially sample the scene at the same rate as the scene dynamics, they offer several advantages over conventional cameras: very high temporal resolution, low latency, very high dynamic range (HDR, \SI{140}{\decibel}) and low power and bandwidth requirements - traits which make them well suited to capturing motion.
Hence, event-based cameras open the door to tackle challenging scenarios that are inaccessible to traditional cameras, such as high-speed and/or HDR tracking~\cite{Mueggler14iros,Zhu17icra,Gallego17pami,Gehrig19ijcv}, control~\cite{Conradt09iscas,Delbruck13fns,Mueller15cdc} and Simultaneous Localization and Mapping (SLAM)~\cite{Kim16eccv,Rebecq17ral,Zhu17cvpr,Rosinol18ral}.
Due to their principle of operation and unconventional output, these cameras represent a paradigm shift in computer vision, and so, new algorithms are needed to unlock their capabilities.
A survey paper on event-based cameras, algorithms, and applications, has been recently published in~\cite{Gallego19arxiv}.

We consider the problem of segmenting a scene viewed by an event-based camera into independently-moving objects.
In the context of traditional cameras, this problem is known as \emph{motion segmentation}~\cite{Zappella08caird}, and it is an essential pre-processing step for several applications in computer vision, such as surveillance, tracking, and recognition~\cite{Ommer09ijcv}.
Its solution consists of analyzing two or more consecutive images from a video camera to infer the motion of objects and their occlusions.
In spite of progress in this field, conventional cameras are not ideally suited to acquiring and analyzing motion; since exposure time is globally applied to all pixels, they suffer from motion blur in fast moving scenes. 
Event-based cameras are a better choice since they sample at exactly the rate of scene dynamics, but conventional techniques cannot be applied to the event data. 

Motion segmentation in the case of a static event-based camera is simple, because in this scenario events are solely due to moving objects (assuming there are no changes in illumination) \cite{Litzenberger06dspws,Ni15neco,Piatkowska12cvprw}.
The challenges arise in the case of a moving camera, since in this scenario events are triggered everywhere on the image plane, produced by both the moving objects as well as the apparent motion of the static scene induced by the camera's ego-motion.
Hence, event-based motion segmentation consists of classifying each event into a different object, including the background.
However, each event carries very little information, 
and therefore it is challenging to perform the mentioned per-event classification.

We propose a method to tackle the event-based motion segmentation problem in its most general case, with a possibly moving camera.
Inspired by classical layered models~\cite{Wang93cvpr}, our method classifies the events of a space-time window 
into separate clusters (i.e., ``layers''), where each cluster represents a coherent moving object (or background) (see Fig.~\ref{fig:method:diagram}).
The method jointly estimates the motion parameters of the clusters and the event-cluster associations (i.e., likelihood of an event belonging to a cluster) in an iterative, alternating fashion, using an objective function based on motion compensation~\cite{Gallego18cvpr,Gallego17ral} 
(basically, the better the estimated unknowns, the sharper the motion-compensated event image of each cluster). 
Our method is flexible, allowing for different types of parametric motions of the objects and the scene (translation, rotation, zooming, etc.).

\vspace{-2ex}
\paragraph{Contributions.}
In summary, our contributions are:
\begin{itemize}[noitemsep,nolistsep]
\itemsep0em 
	\item A novel, iterative method for segmenting multiple objects based on their apparent motion on the image plane, producing a per-event classification into space-time clusters described by parametric motion models.
    \item The detection of independently moving objects without having to compute optical flow explicitly. 
    Thus, we circumvent this difficult and error-prone step toward reaching the goal of motion-based segmentation.
	\item A thorough evaluation in challenging, real-world scenarios, such as high-speed and difficult illumination conditions, which are inaccessible to traditional cameras (due to severe motion blur and HDR), outperforming the state-of-the-art by as much as \SI{10}{\percent}, and showing that accuracy in resolving small motion differences between objects is a central property of our method.
\end{itemize}
As a by-product, our method produces sharp, motion-compensated images of warped events, which represent the appearance (i.e., shape or edge-map) of the segmented objects (or background) in the scene (Fig.~\ref{fig:method:diagram}, Right).

The rest of the paper is organized as follows:
Section~\ref{sec:related_work} reviews related work on event-based motion segmentation, 
Section~\ref{sec:method} describes the proposed solution, which is then evaluated in Section~\ref{sec:experiments}.
Conclusions are drawn in Section~\ref{sec:conclusion}.

\section{Related Work}
\label{sec:related_work}

Event-based motion segmentation in its non-trivial form (i.e., in the presence of event-clutter caused by camera ego-motion, or a scene with many independently moving, overlapping objects) has been addressed before~\cite{Glover16iros,Vasco17icar,Stoffregen17acra,Mitrokhin18iros,Barranco15iccv}.

In \cite{Glover16iros}, a method is presented for detection and tracking of a circle in the presence of event clutter.
It is based on the Hough transform using optical flow information extracted from temporal windows of events.
Segmentation of a moving object in clutter was also addressed in~\cite{Vasco17icar}.
It considered more generic object types than~\cite{Glover16iros} by using event corners as primitives, 
and it adopted a learning technique to separate events caused by camera motion from those due to the object. 
However, the method required the additional knowledge of the robot joints controlling the camera.

Segmentation has been recently addressed by \cite{Stoffregen17acra,Mitrokhin18iros} using the idea of motion-compensated event images~\cite{Gallego18cvpr,Gallego17ral,Gallego19cvpr,Stoffregen19cvpr,Zhu19cvpr,Dardelet18arxiv}.
For example, \cite{Stoffregen17acra} first fitted a motion compensation model to the dominant events, then removed these and fitted another motion model to the remaining events, greedily.
Similarly, \cite{Mitrokhin18iros} detected moving objects in clutter by fitting a motion-compensation model to the dominant events (i.e., the background) and detecting inconsistencies with respect to that motion (i.e., the objects).
The objects were then ``segmented'' via morphological operations on the warped image, and were used for tracking. 
The method could handle occlusions during tracking, but not during detection.

Our method differs from~\cite{Glover16iros} in that we demonstrate segmentation on objects with arbitrary shapes, and from~\cite{Vasco17icar} in that we do not require additional inputs (e.g., robot joints).
Our work is most related to~\cite{Stoffregen17acra,Mitrokhin18iros}; however, it has the following novelties:
(\emph{i}) it actually performs \textit{per-event} segmentation, rather than just providing bounding boxes for detected object regions,
(\emph{ii}) it allows for general parametric motion models (as those in~\cite{Gallego18cvpr}) to describe each cluster,
(\emph{iii}) it performs optimization based on a single objective function (as opposed to two sequential optimization criteria in~\cite{Mitrokhin18iros}),
(\emph{iv}) it is able to handle occlusions between objects at any point in time.
The proposed method, is, to our knowledge, the first one that \emph{jointly} estimates the apparent motion parameters of all objects in the scene and the event-object associations (i.e., segmentation).
It does so by leveraging the idea of motion-compensated event images within an iterative, alternating optimization approach, in an expectation-maximization (EM) fashion.

\section{Methodology}
\label{sec:method}

Our method is inspired by the combination of classical layered models~\cite{Wang93cvpr} and event-based motion compensation~\cite{Gallego18cvpr}.
In the following, we review the working principle of event-based cameras, describe the motion segmentation problem addressed and present our proposed solution.
\global\long\def\walkingwidth{0.48\linewidth}
\begin{figure}[t]
    \vspace{-1ex}
    \centering
    \subfloat[Cluster 1]{\frame{\includegraphics[width=\walkingwidth]{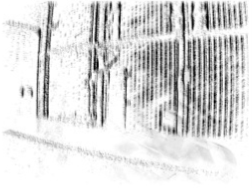}}\label{fig:walk:background}}~~
    \subfloat[Cluster 2]{\frame{\includegraphics[width=\walkingwidth]{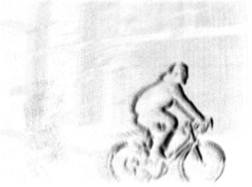}}\label{fig:walk:bike}}\\[-1ex]
    \subfloat[Cluster 3]{\frame{\includegraphics[width=\walkingwidth]{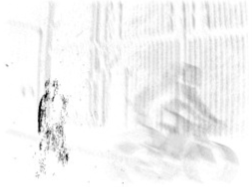}}\label{fig:walk:person}}~~
    \subfloat[All clusters (merged IWEs)]{\frame{\includegraphics[width=\walkingwidth]{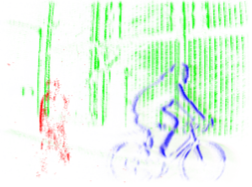}}\label{fig:walk:global}}
    \caption{\label{fig:method:layers}%
	Our method splits the events into clusters (one per moving object), producing motion-compensated images of warped events (IWEs), as shown in (a)-(c) for the three objects in the scene of Fig.~\ref{fig:method:diagram}.
The likelihood of each event is represented by the darkness of the pixel. 
	Since the likelihoods are nonzero, ``ghosts'' can be seen in the individual clusters.
	IWEs in (a)-(c) are merged into a single image (d), using a different color for each cluster.
	Segmented events in upcoming experiments are shown using this colored motion-compensated image representation.}
	\vspace{-2.75ex}
\end{figure}

\vspace{-2ex}
\paragraph{Event-Based Camera Working Principle.}
Event-based cameras, such as the DVS~\cite{Lichtsteiner08ssc}, have independent pixels that output ``events'' in response to intensity changes. 
Specifically, if $L(\bx,t)\doteq \log I(\bx,t)$ is the logarithmic intensity at pixel $\bx \doteq (x,y)^\top$ on the image plane, the DVS generates an event $e_k \doteq (\bx_k,t_k,\pol_k)$ 
if the change in intensity at pixel $\bx_k$ reaches a threshold $C$ (e.g., 10-15\% relative change):
\vspace{-0.5ex}
\begin{equation}
\label{eq:generativeEventCondition}
\Delta L(\bx_k,t_k) \doteq L(\bx_k,t_k) - L(\bx_k,t_k-\Delta t_k) = \pol_{k}\, C,
\vspace{-0.5ex}
\end{equation}
where $t_k$ is the timestamp of the event, 
$\Delta t_k$ is the time since the previous event at the same pixel $\bx_k$ 
and $\pol_k \in \{+1,-1\}$ is the polarity of the event (the sign of the intensity change).

\subsection{Problem Statement}
Since each event carries little information and we do not assume prior knowledge of the scene, we process events in packets (or groups) to aggregate sufficient information for estimation.
Specifically, given a packet of events $\cE \doteq \{e_k\}_{k=1}^{\numEvents}$ in a space-time volume of the image plane $V\doteq \Omega\times T$, we address the problem of classifying them into $\numLayers$ clusters (also called ``layers''), with each cluster representing a coherent motion, of parameters $\bparams_j$.
We assume that $T$ is small enough so that the motion parameters of the clusters $\bparams \doteq \{\bparams_j\}_{j=1}^{\numLayers}$ are constant.

The images on both sides of the algorithm block in Fig.~\ref{fig:method:diagram} illustrate the above-mentioned problem and its solution, respectively.
Notice that, ($i$) since events have space-time coordinates, clusters are three-dimensional, contained in $V$,
and ($ii$) since corresponding events (caused by the same point of a moving edge) describe point trajectories in $V$, optimal clusters should contain them, therefore, clusters have a ``tubular'' shape (Fig.~\ref{fig:method:diagram}, segmented events).
Implicit in motion segmentation, if two objects share the same motion, they are segmented together, regardless of their location.

\subsection{Summary of Proposed Solution}
\label{sec:method:summary}
Leveraging the idea of motion compensation~\cite{Gallego18cvpr}, we seek to separate the events $\cE$ into clusters by maximizing \emph{event alignment}, i.e., maximizing the sharpness of motion-compensated images (one per cluster) of warped events.

More specifically, the idea of motion compensation~\cite{Gallego18cvpr} is that, as an edge moves on the image plane, it triggers events on the pixels it traverses.
The motion of the edge can be estimated by warping the events to a reference time and maximizing their alignment, producing a sharp Image of Warped Events (IWE)~\cite{Gallego18cvpr}.
In the case of multiple objects with different motions, maximal event alignment cannot be achieved using a single warp, and so, several warps (i.e., motion models or ``clusters'') are required, as well as identifying which events belong to which object (i.e., ``event-cluster associations''). 
This is the essence of our approach, which is illustrated in Figs.~\ref{fig:method:diagram} and~\ref{fig:method:layers}. 
Fig.~\ref{fig:method:diagram} shows the events produced by three objects in a scene: a pedestrian, a cyclist and a the building facade (camera motion).
Each object has a different motion 
and triggers events on the image plane as it moves. 
When events are warped to a reference time (e.g., $\tref=0$) according to a candidate motion model, 
they produce an IWE.
If the candidate motion coincides with the true motion of the object causing the events, the warped events align, producing a sharp motion-compensated IWE, as shown in Fig.~\ref{fig:method:layers} 
using three different motion models (one per object).
Otherwise, they do not align, producing a blurred IWE.
We use the sharpness of such IWE as the main cue to segment the events.
Our method jointly identifies the events corresponding to each independently moving object as well as the object's motion parameters.

\subsection{Mathematical Formulation}
In contrast to previous methods~\cite{Stoffregen17acra,Mitrokhin18iros}, we explicitly model event-cluster associations in the motion-compensation framework, i.e., $p_{kj} = P(e_k \in \layer_j)$ is the probability of the $k$-th event belonging to the $j$-th cluster.
Let $\probmat \equiv (p_{kj})$ be an $\numEvents \times \numLayers$ matrix with all event-cluster associations.
The entries of $\probmat$ must be non-negative, and each row must add up to one.
Using these associations, we define the \emph{weighted} IWE of the $j$-th cluster as
\vspace{-0.5ex}
\begin{equation}
\IWE_{j}(\bx) \doteq \textstyle{\sum_{k=1}^{\numEvents}}\,p_{kj}\,\delta(\bx-\bx'_{kj}),
\label{eq:weightedIWE}
\end{equation}
with $\bx'_{kj} = \Warp(\bx_{k},t_{k};\bparams_j)$ the warped event location, and $\delta$ the Dirac delta. 
Equation~\eqref{eq:weightedIWE} states that events are warped, 
\vspace{-0.5ex}
\begin{equation}
e_k\doteq (\bx_k,t_k,\pol_k)\mapsto e'_k \doteq (\bx'_k,\tref,\pol_k),
\vspace{-0.5ex}
\end{equation}
and the values $p_{kj}\geq 0$ (i.e., weights) are accumulated at the warped locations $\bx'_k$.
Event alignment within the $j$-th cluster is measured using image contrast~\cite{Gonzalez09book}, 
which is defined by a sharpness/dispersion metric, such as the variance~\cite{Gallego18cvpr}:
\vspace{-0.5ex}
\begin{equation}
\label{eq:ContrastOneLayer}
\variance(\IWE_j) \doteq \frac{1}{|\Omega|}\int_{\Omega} (\IWE_{j}(\bx)-\mu_{\IWE_{j}})^2 d\bx,
\vspace{-0.5ex}
\end{equation}
where $\mu_{\IWE_{j}}$ is the mean of the IWE over the image plane $\Omega$.

We propose to find the associations $\probmat$ and cluster parameters $\bparams$ that maximize the sum of contrasts of all clusters:
\vspace{-0.5ex}
\begin{equation}
\label{eq:weightedContrast}
(\bparams^\ast,\probmat^\ast) = \argmax_{(\bparams,\probmat)} \sum_{j=1}^{\numLayers}\variance(\IWE_{j}).
\vspace{-0.5ex}
\end{equation}
Since the problem addressed does not admit a closed-form solution, we devise an iterative, alternating optimization approach, which we describe in the next section.

The pseudo-code of our method is given in Algorithm~\ref{alg:ems}.
From the output of Algorithm~\ref{alg:ems}, it is easy to compute motion-compensated images of events corresponding to each cluster, i.e., the weighted IWEs~\eqref{eq:weightedIWE} shown in Fig.~\ref{fig:method:layers}.
Each IWE shows the sharp, recovered edge patterns (i.e, and appearance model) of the objects causing the events.
\begin{algorithm}[t!]
\caption{Event-based Motion Segmentation}
\label{alg:ems}
\begin{algorithmic}[1]
\State \textbf{Input}: events $\cE=\{e_k\}_{k=1}^{\numEvents}$ in a space-time volume $V$ of the image plane, and number of clusters $\numLayers$.
\State \textbf{Output}: cluster parameters $\bparams=\{\bparams_j\}_{j=1}^{\numLayers}$ and event-cluster assignments $\probmat \equiv p_{kj} \doteq P(e_k \in \layer_j)$.
\State \textbf{Procedure:}
\State Initialize the unknowns $(\bparams,\probmat)$ (see Section~\ref{sec:method:initialization}).\label{alg:line:init}
\State \textbf{Iterate} until convergence:\label{alg:line:loop}
  \State $\bullet$ Compute the event-cluster assignments $p_{kj}$ using~\eqref{eq:probEventInLayer}.\label{alg:line:assign}
  \State $\bullet$ Update the motion parameters of all clusters~\eqref{eq:argmaxLayerParams}.\label{alg:line:mot}
\end{algorithmic}
\end{algorithm}

\subsection{Alternating Optimization}
\label{sec:method:alternating}
Each iteration of Algorithm~\ref{alg:ems} has two steps (lines~\ref{alg:line:assign} and~\ref{alg:line:mot}), 
as in a coordinate ascent algorithm.
If the associations $\probmat$ are fixed, we may update the motion parameters
\vspace{-0.5ex}
\begin{equation}
\label{eq:argmaxLayerParams}
\bparams \leftarrow \bparams + \mu \nabla_{\bparams} \Bigl( \textstyle{\sum_{j=1}^{\numLayers}}\variance(\IWE_j) \Bigr)
\vspace{-0.5ex}
\end{equation}
by taking a step ($\mu \ge 0$) in an ascent direction of the objective function~\eqref{eq:weightedContrast} with respect to the motion parameters.
Motion-compensation methods~\cite{Gallego18cvpr,Mitrokhin18iros} typically use gradient ascent or line search to solve for the motion parameters that maximize some objective function of the IWE.
In our case, because the IWE~\eqref{eq:weightedIWE} depends on both $\bparams$ and $\probmat$ and we seek to jointly estimate them, we do not wastefully search for the best $\bparams$ given the current estimate of $\probmat$.
Instead, we update $\bparams$ using \eqref{eq:argmaxLayerParams}, proceed to refine $\probmat$ (see~\eqref{eq:probEventInLayer}), and iterate.

Fixing the motion parameters $\bparams$,
we may refine the associations $\probmat$ using a closed-form probability partitioning law:
\vspace{-0.5ex}
\begin{equation}
\label{eq:probEventInLayer}
p_{kj} = \frac{c_j(\bx'_{k}(\bparams_{j}))}{\sum_{i=1}^{\numLayers}c_i(\bx'_{k}(\bparams_{i}))},
\vspace{-0.5ex}
\end{equation}
where $c_j(\bx)\neq 0$ is the local contrast (i.e., sharpness) of the $j$-th cluster at pixel $\bx$, 
and it is given by the value of the weighted IWE, $c_j(\bx) \doteq \IWE_j(\bx)$.
Thus, each event is softly assigned to each cluster based on how it contributes to the sharpness of all $\numLayers$ IWEs.
The alternating optimization approach in Algorithm~\ref{alg:ems} resembles the EM algorithm, with the E-step given by~\eqref{eq:probEventInLayer} and the M-step given by~\eqref{eq:argmaxLayerParams}.

\subsection{Initialization}
\label{sec:method:initialization}
The proposed alternating method converges locally (i.e., there is no guarantee of convergence to a global solution), and it requires initialization of $\bparams,\probmat$
to start the iteration.

Several initialization schemes are possible, depending on the motion models.
For example, if the warps of all clusters are of optical flow type, one could first extract optical flow from the events (e.g., using~\cite{Benosman14tnnls,Zhu18rss}) and then cluster the optical flow vectors (e.g., using the k-means algorithm). 
The resulting cluster centers in velocity space would provide an initialization for the motion parameters of the clusters $\bparams$.

We follow a greedy approach, similar to that in~\cite{Stoffregen17acra}, 
that works well in practice, providing cluster parameters close to the desired ones. 
It is valid regardless of the motion models used. 
We initialize events to have equal association probabilities, and then maximize the contrast of the first cluster with respect to its motion parameters. 
We then find the gradient of the local contrast for each event with respect to the motion parameters. 
Those events that belong to the cluster under consideration become less ``in focus'' when we move away from the optimized parameters, so those events that have a negative gradient are given a high association probability for that cluster and a low one for clusters subsequent. 
The process is repeated for the remaining clusters until all motion parameters $\bparams$ and associations $\probmat$ have been filled.

\subsection{Discussion of the Segmentation Approach}
\label{sec:method:discussion}

The proposed approach is versatile, since it allows us to consider diverse parametric motion/warping models, such as linear motion 
(optic flow)~\cite{Zhu17icra,Stoffregen17acra,Gallego18cvpr},
rotational motion~\cite{Gallego17ral}, 4-DOF (degrees-of-freedom) motion~\cite{Mitrokhin18iros}, 8-DOF homographic motion~\cite{Gallego18cvpr}, etc. 
Moreover, each cluster may have a different motion model, $\{\Warp_{j}\}_{j=1}^{\numLayers}$, as opposed to having a single model for all events, and therefore, all clusters.
This characteristic is unique of our method.

It is also worth noting that the proposed method classifies events according to motion without having to explicitly compute optical flow, which is a widespread motion descriptor. 
Thus, our method is not simply optical flow clustering.
Instead, our method encapsulates motion information in the warps 
of each cluster, thus by-passing the error-prone step of optical flow estimation in favor of achieving the desired goal of motion segmentation of the events.

The edge-like motion-compensated IWEs corresponding to each cluster are, upon convergence, a description of the intensity patterns (entangled with the motion) that caused the events.
Thus our method recovers fine details of the appearance (e.g., shape) of the objects causing the events without having to estimate a (computationally expensive) 3D scene representation. 
In~\cite{Mitrokhin18iros} fine details were only available for the dominant motion cluster.

Finally, the number of clusters $\numLayers$ is a hyper-parameter that may be tuned by a meta-algorithm (in the experiments, we set $\numLayers$ manually). 
This is a well-known topic in clustering~\cite{Fraley98comjnl}.
While automatically determining the optimal $\numLayers$ depending on the scene is outside the scope of this paper, 
it should be noted that as we show in Section \ref{sec:exp:num_clusters}, our method is not sensitive to excess clusters $\numLayers$.

\subsection{Sequence Processing}
\label{sec:method:sequence}
The above method segments the events $\cE$ from a short time interval $T$.
To process an entire stream of events, a sliding window approach is used, splitting the stream into packets of events $\{\cE_{n}\}_{n=1}^{N_g}$.
We process the \mbox{$n$-th} packet and then slide the window, thus selecting more recent events.
The motions estimated by clustering $\cE_{n}$ can be propagated in time to predict an initialization for the clusters of the next packet, $\cE_{n+1}$.
We use a fixed number of events $\numEvents$ per window, and slide by half of it, $\numEvents/2$.
\vspace{-1ex}

\section{Experiments}
\label{sec:experiments}

\global\long\def\seqfast{{\small\texttt{Fast Moving Drone}}}
\global\long\def\seqmultobj{{\small\texttt{Multiple Objects}}}
\global\long\def\seqlightvar{{\small\texttt{Lightning variation}}}
\global\long\def\seqwib{{\small\texttt{What is Background?}}}
\global\long\def\seqoccl{{\small\texttt{Occluded sequence}}}

\paragraph*{Overview.}
In this section we first provide a quantitative evaluation of our method on a publicly available, real-world dataset \cite{Mitrokhin18iros}, showing that we significantly outperform two baseline methods~\cite{Stoffregen17acra,Mitrokhin18iros}.
We provide further quantitative results on the accuracy of our method with regard to \emph{relative motion differences}
and we demonstrate the efficacy of our method on additional, challenging real-world data.
Throughout the experiments, we demonstrate several features of our method, namely that
(\emph{i}) it allows arbitrary motion models for different clusters,
(\emph{ii}) it allows us to perform motion segmentation on difficult scenes (high speed and/or HDR), where conventional cameras would fail,
(\emph{iii}) it is robust to the number of clusters used ($\numLayers$),
and (\emph{iv}) that it is able to perform motion segmentation on non-rigid scenes.
The sequences considered cover a broad range of motion speeds, from \SI{12}{pixel\per\second} to several hundred \si{pixel\per\second}.

We strongly recommend looking at the accompanying video and supplementary material, where we present further experiments, including a comparison to ``naive'' k-means clustering, mixture density models and fuzzy-k-means.

The following experiments were carried out with data from a DAVIS240C camera~\cite{Brandli14ssc}, which provide both events and grayscale frames. 
The frames are not used in the experiments; they serve only an illustrative purpose.

\subsection{Quantitative Evaluation}

\begin{figure}[t]
\global\long\def\widthscenesmitr{0.3\linewidth}
    \centering
    \subfloat[]{\label{fig:scenes:Mitrokhin_mo}%
    \frame{\includegraphics[width=\widthscenesmitr]{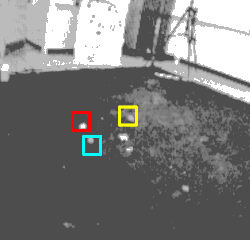}}}~
    \subfloat[]{\label{fig:scenes:Mitrokhin_os}%
    \frame{\includegraphics[width=\widthscenesmitr]{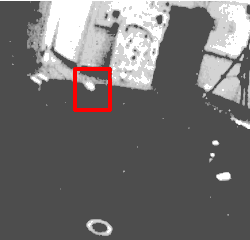}}}~
    \subfloat[]{\label{fig:scenes:Mitrokhin_wib}%
    \frame{\includegraphics[width=\widthscenesmitr]{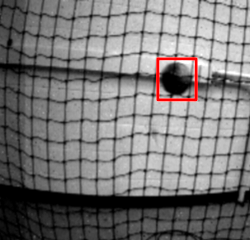}}}
    \vspace{-1.5ex}
    \caption{Several scenes from the Extreme Event Dataset (EED) \cite{Mitrokhin18iros}:
    (a) {\small\texttt{Multiple Objects}}, 
    (b) {\small\texttt{Occluded Sequence}} and (c) {\small\texttt{What is Background?}}
    Moving objects (drones, balls, etc.) are within hand-labeled bounding boxes.
    Images have been brightened for visualization.\label{fig:scenes:Mitrokhin}}
    \vspace{3ex}
    \begin{adjustbox}{max width=\linewidth}
    \setlength{\tabcolsep}{2pt}
    \begin{tabular}{lccc}
    \toprule
        \textbf{EED Sequence Name} & \textbf{SOFAS}~\cite{Stoffregen17acra} &\textbf{Mitrokhin}~\cite{Mitrokhin18iros} & \textbf{Ours (Alg.~\ref{alg:ems})} \\ \midrule
        {\small\texttt{Fast moving drone}} & 88.89 & 92.78 &  \textbf{96.30}\\
        {\small\texttt{Multiple objects}} & 46.15 & 87.32 &  \textbf{96.77}\\
        {\small\texttt{Lighting variation}} & 0.00 & \textbf{84.52} &  80.51\\
        {\small\texttt{What is Background?}} & 22.08 & 89.21 &  \textbf{100.00}\\
        {\small\texttt{Occluded sequence}} & 80.00 & 90.83 &  \textbf{92.31}\\ \bottomrule
    \end{tabular}
    \end{adjustbox}
    \vspace{-1ex}
    \captionof{table}{\label{tab:iros18}%
    Comparison with state-of-the-art using the success rate proposed by~\cite{Mitrokhin18iros} of detection of moving objects (in~\si{\percent}).}
    \vspace{-2ex}
\end{figure}

\paragraph{Results on Dataset from~\cite{Mitrokhin18iros}.}
We ran our segmentation method on the Extreme Event Dataset (EED) from~\cite{Mitrokhin18iros} and compared against the results from \cite{Mitrokhin18iros} and \cite{Stoffregen17acra}.
The sequences in the EED dataset showcase a range of scenes (Fig.~\ref{fig:scenes:Mitrokhin} and Table~\ref{tab:iros18}) which are very challenging for conventional cameras. 
In particular, they comprise fast moving objects (around \SI{600}{pixel/\second}) in \seqfast\; and \seqmultobj, which are almost indiscernible on the frames provided by the DAVIS camera, 
as well as scenes with extreme lighting variations, such as \seqlightvar\, (in which a drone is tracked despite a strobe light pointing at the camera),
and object occlusions.
Having object segmentation rather per-event segmentation in mind,
the EED dataset provides timestamped bounding boxes around the moving objects in the scene 
and proposes to measure \emph{object segmentation success} whenever the estimated bounding box overlaps at least \SI{50}{\percent} with the hand-labeled one \emph{and} it has more area within the hand-labeled bounding box than outside.
To compare against~\cite{Mitrokhin18iros}, we perform motion segmentation on the events that occur around the timestamp of the bounding-box and count success if for a given cluster the above criterion is true for the segmented events.
For a fair comparison, we used the same type of motion models (4-DOF warps) as in~\cite{Mitrokhin18iros}.

Table~\ref{tab:iros18} reports the results of the comparison of our method against \cite{Stoffregen17acra} and \cite{Mitrokhin18iros}.
Our method outperforms \cite{Stoffregen17acra} in all sequences by a large margin (from \SIrange{7.41}{84.52}{\percent}), 
and improves over \cite{Mitrokhin18iros} in all but one sequence, where it has comparable performance. 
In four out of five sequences we achieve accuracy above \SI{92}{\percent}, and in one of them, a perfect score, \SI{100}{\percent}.
Some results of the segmentation are displayed on the first columns of Fig.~\ref{fig:allseqs}.
In the \seqwib\, scene (1st column of Fig.~\ref{fig:allseqs}), a ball is thrown from right to left behind a net while the camera is panning, following the ball. 
Our method clearly segments the scene into two clusters: the ball and the net, correctly handling occlusions.
In the \seqlightvar\, (2nd column of Fig.~\ref{fig:allseqs}), a quadrotor flies through a poorly lit room with strobe lightning in the background, 
and our method is able to segment the events due to the camera motion (green and blue) and due to the quadrotor (purple).

\vspace{-2ex}
\paragraph{Accuracy vs Displacement.}
While the dataset from \cite{Mitrokhin18iros} provides a benchmark for comparison against the state-of-the-art, it does not allow us to assess the \emph{per-event} accuracy of our method.
Here we measure segmentation success directly as a percentage of correctly classified events, thus much more fine-grained than with bounding boxes. 
Since real data contains a significant proportion of noise events, we perform the quantitative analysis on event data from a state-of-the-art photorealistic simulator~\cite{Rebecq18corl}. 
Knowing which objects generated which events, allows us to finely resolve the accuracy of our method.
\global\long\def\layerwidth{0.2\linewidth}
\begin{figure*}[t]
	\centering
	\begin{adjustbox}{max width=\textwidth}
	\bgroup
    \def\arraystretch{1}
    \setlength{\tabcolsep}{2pt}
    {\normalsize
	\begin{tabular}{
	>{\centering\arraybackslash}m{\layerwidth} 
	>{\centering\arraybackslash}m{\layerwidth}
	>{\centering\arraybackslash}m{\layerwidth} 
	>{\centering\arraybackslash}m{\layerwidth}
	>{\centering\arraybackslash}m{\layerwidth}
	>{\centering\arraybackslash}m{\layerwidth}}
		Ball behind net~\cite{Mitrokhin18iros}
		&
		Drone, low light~\cite{Mitrokhin18iros}
		&
		Traffic scene
		&
		Buildings and car
		&
		Street, facing the sun
		&		
		Fan and coin		
		\\
		\ifcompileanimatedexperimentpage
		\frame{\animategraphics[width=\linewidth]{15}{images/wib/img_00}{04}{30}}
	    \else
	    \frame{\includegraphics[width=\linewidth]{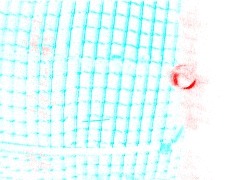}}
	    \fi
		&
		\ifcompileanimatedexperimentpage
	    \frame{\animategraphics[width=\linewidth]{15}{images/hdr/segmented/segmented-}{0}{45}}
	    \else
		\frame{\includegraphics[width=\linewidth]{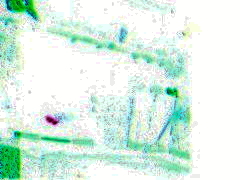}}
		\fi
		&
		\ifcompileanimatedexperimentpage
	    \frame{\animategraphics[width=\linewidth]{15}{images/mensa/recolor-}{0}{116}}
	    \else
		\frame{\includegraphics[width=\linewidth]{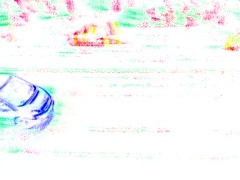}}
		\fi
		&
		\ifcompileanimatedexperimentpage
	    \frame{\animategraphics[width=\linewidth]{15}{images/car/image_00}{10}{95}}
	    \else
		\frame{\includegraphics[width=\linewidth]{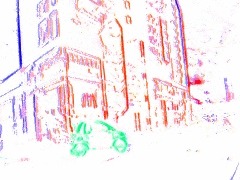}}
		\fi
		&
		\ifcompileanimatedexperimentpage
    	\frame{\animategraphics[width=\linewidth]{15}{images/hdrb/hdr_bike_0}{057}{210}}
	    \else
		\frame{\includegraphics[width=\linewidth]{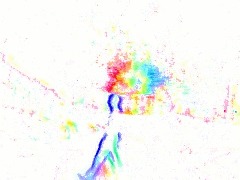}}
		\fi
		&
		\ifcompileanimatedexperimentpage
	    \frame{\animategraphics[width=\linewidth]{12}{images/fan/output-}{0}{53}}
	    \else
		\frame{\includegraphics[width=\linewidth]{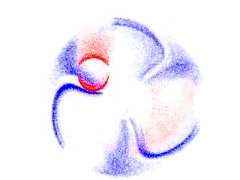}}
		\fi
		\\
		\frame{\includegraphics[width=\linewidth]{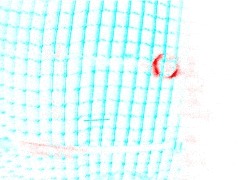}}
		&
		\frame{\includegraphics[width=\linewidth]{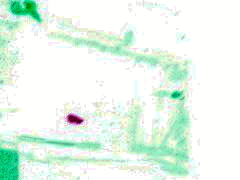}}
		&
		\frame{\includegraphics[width=\linewidth]{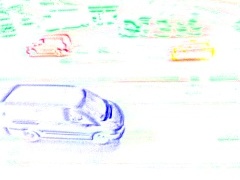}}
		&
		\frame{\includegraphics[width=\linewidth]{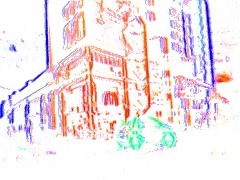}}
		&
		\frame{\includegraphics[width=\linewidth]{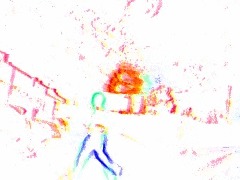}}
		&
		\frame{\includegraphics[width=\linewidth]{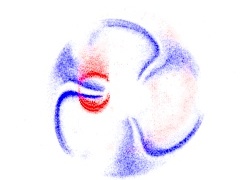}}
		\\
		\frame{\includegraphics[width=\linewidth]{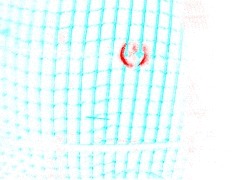}}
		&
		\frame{\includegraphics[width=\linewidth]{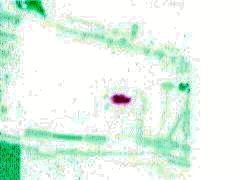}}
		&
		\frame{\includegraphics[width=\linewidth]{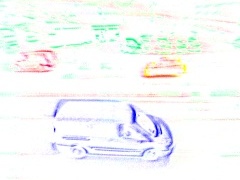}}
		&
		\frame{\includegraphics[width=\linewidth]{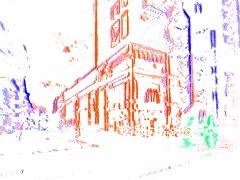}}
		&
		\frame{\includegraphics[width=\linewidth]{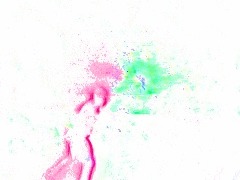}}
		&
		\frame{\includegraphics[width=\linewidth]{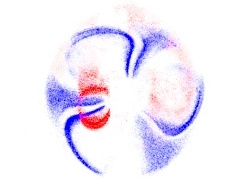}}
		\\
		\frame{\includegraphics[width=\linewidth]{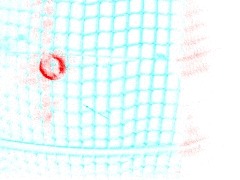}}
		&
		\frame{\includegraphics[width=\linewidth]{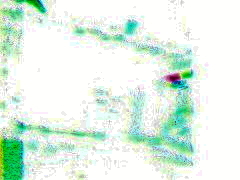}}
		&
		\frame{\includegraphics[width=\linewidth]{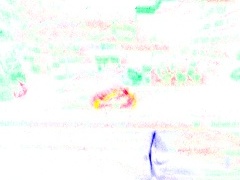}}
		&
		\frame{\includegraphics[width=\linewidth]{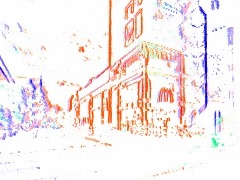}}
		&
		\frame{\includegraphics[width=\linewidth]{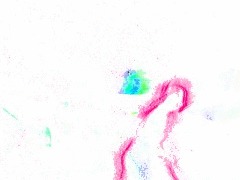}}
		&
		\frame{\includegraphics[width=\linewidth]{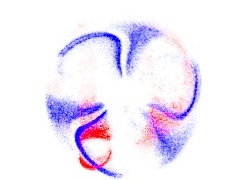}}
	\end{tabular}
	}
	\egroup
	\end{adjustbox}
	\vspace{-1ex}
	\caption{From top to bottom: snapshots (motion-compensated images, as in Fig.2) of events segmented into clusters on multiple sequences (one per column).
    Events colored by cluster membership.
    Best viewed in the accompanying video.}
    \vspace{-1ex}
	\label{fig:allseqs}
\end{figure*}

\global\long\def\heightrocksmetric{2.8cm}
\begin{figure}[t]
    \centering%
    \subfloat[Events (red and blue).]{%
    \includegraphics[height=\heightrocksmetric]{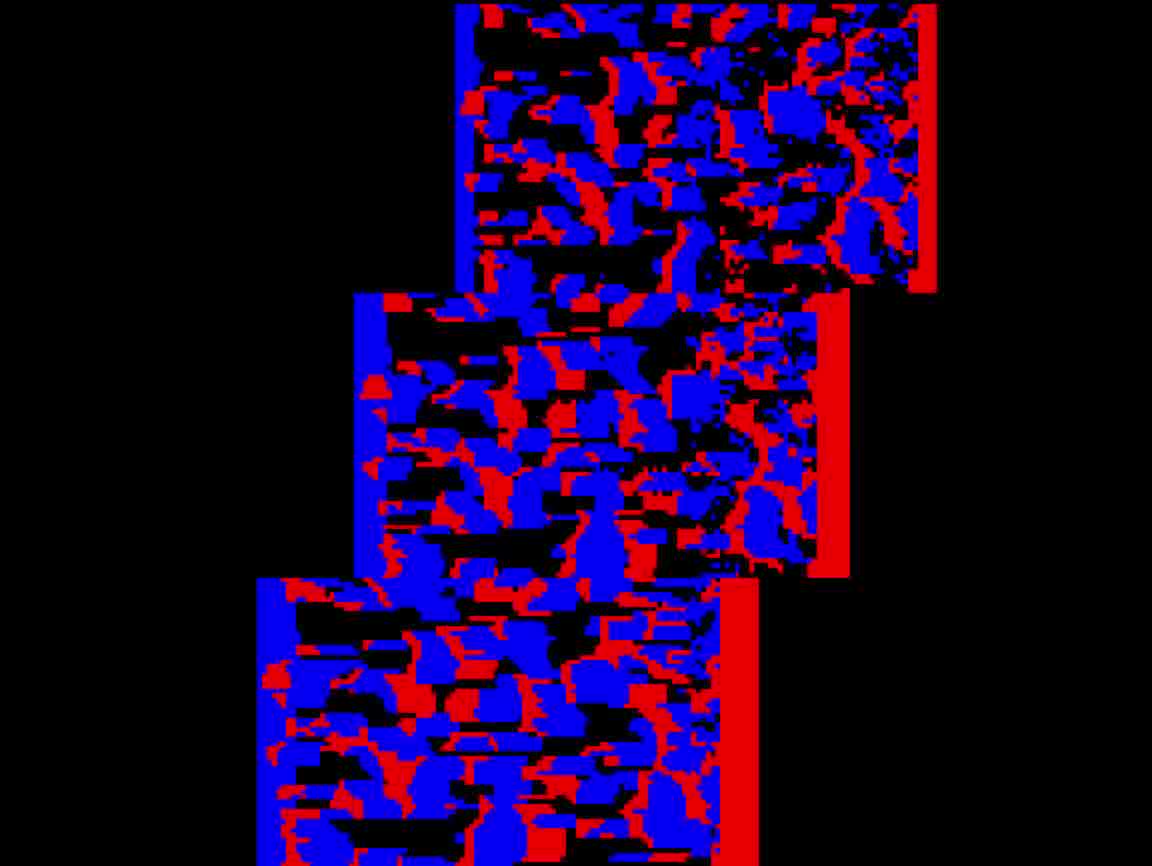}
    \label{fig:metric:scene}}~
    \subfloat[Success Rate.]{%
		\includegraphics[height=\heightrocksmetric]{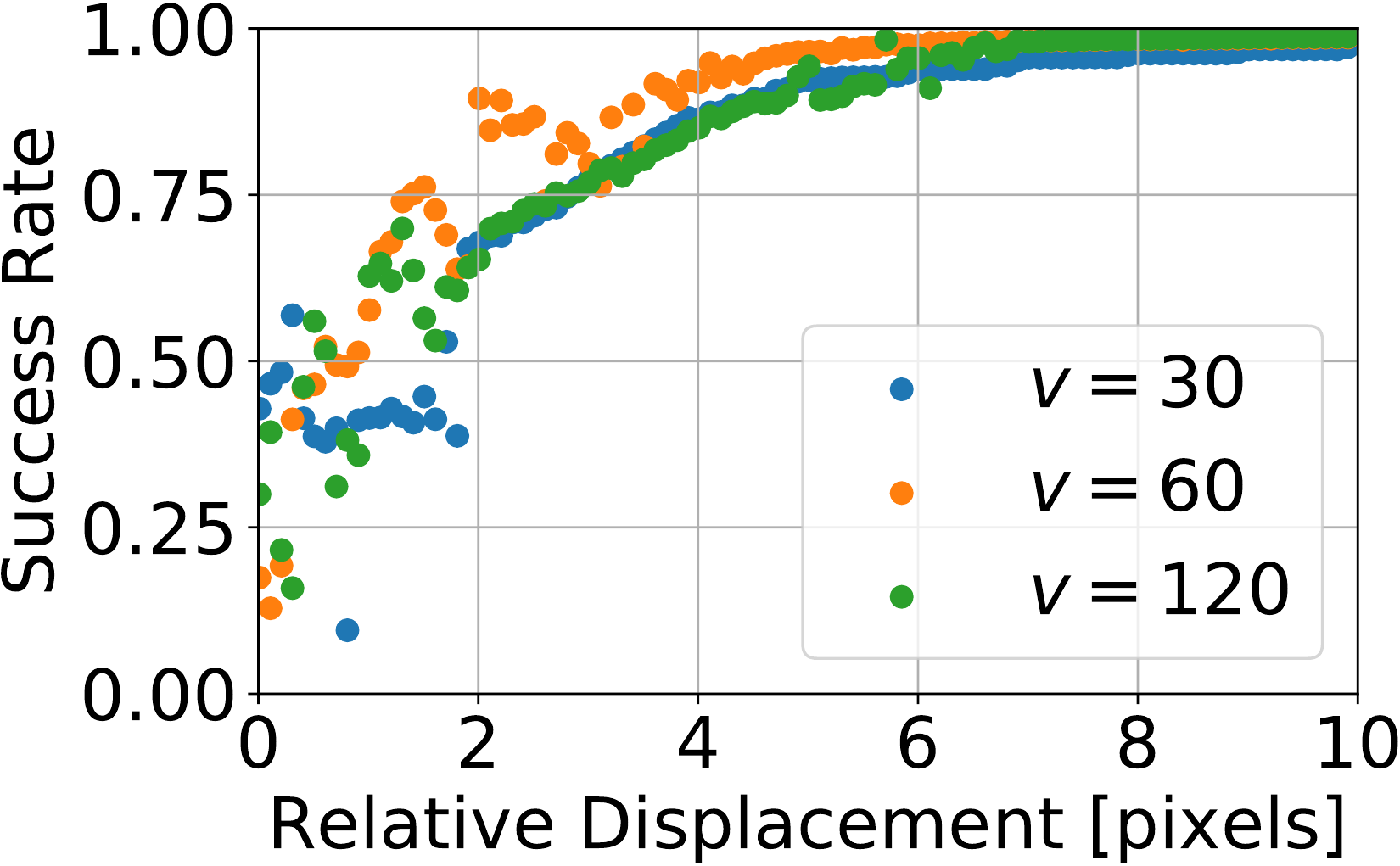}
    \label{fig:metric:curves}}
    \vspace{-1ex}
    \caption{\label{fig:metric}
    \emph{Per-event Success Evaluation.}
Segmentation accuracy vs relative object displacement (\ref{fig:metric:curves}) on {\small\texttt{pebbles}} sequence (\ref{fig:metric:scene}) at various relative velocities of objects $v_{\text{rel}}=\{30, 60, 120\}\si{pixel/\second}$).}
    \vspace{-2ex}
\end{figure}

However, segmentation accuracy is closely coupled with the observation window over which events are collected. 
Intuitively, this makes sense; observing two objects with a relative velocity of \SI{1}{pixel/\second} for only \SI{1}{\second} means that the objects have moved only \SI{0.1}{pixels} relative to each other, a difference that is difficult to measure. 
Observing these two objects for \SI{10}{\second} means a relative displacement of \SI{10}{pixels}, which is easier to distinguish.

Fig~\ref{fig:metric} evaluates the above effect on a sequence consisting of textured pebbles moving with different relative velocities (Fig.~\ref{fig:metric:scene}, with events colored in red and blue, according to polarity).
The plot on Fig~\ref{fig:metric:curves} shows that as the relative displacement increases, the proportion of correctly classified events, and therefore, the segmentation accuracy, increases. 
Our method requires that roughly \SI{4}{pixels} of relative displacement have occurred in order to achieve \SI{90}{\percent} accuracy.
This holds true for any relative velocity.
\vspace{-1ex}
\paragraph{Computational Performance.}
The complexity of Algorithm~\ref{alg:ems} is linear in the number of clusters, events, pixels of the IWE and the number of optimization iterations, in total, $O((\numEvents+\numPixels)\numLayers\numIters)$. 
Our method generally converges in less than ten iterations of the algorithm, although this clearly depends on several parameters, such as the data processed.
Further details are given in the supplementary material.
Here, we provide a ballpark figure for the processing speed.
We ran our method on a single core, \SI{2.4}{\giga\Hz} CPU where we got a throughput of \SI{240000}{events/\second} for optical-flow--type warps (Table~\ref{tab:CPU_vs_GPU}).
Almost \SI{99}{\percent} of the time was spent in warping events, which is parallelizable.
Using a GeForce 1080 GPU, we achieved a $10\times$ speed-up factor, as reported in Table~\ref{tab:CPU_vs_GPU}.
The bottleneck is not in computation but rather in memory transfer to and from the GPU.

Throughput decreases as $\numLayers$ increases, since all of the events need to be warped for every extra cluster in order to generate motion compensated images. Further, extra clusters add to the dimensionality of the optimization problem that is solved during the motion-compensation step.

\begin{table}[h!]
    \centering
    \begin{adjustbox}{max width=.7\linewidth}
    \begin{tabular}{SSS}
    \toprule
        $\boldsymbol{\numLayers}$ & \textbf{CPU} [\si{kevents/s}] & \textbf{GPU} [\si{kevents/s}] \\ \midrule
       2    &  239.86 &  3963.20  \\
       5    &  178.19 &  1434.66  \\
       10   &   80.93 &   645.02  \\
       20   &   32.43 &   331.50  \\ 
	   50	&   12.62 &   113.78  \\ \bottomrule
    \end{tabular}
    \end{adjustbox}
    \vspace{-1ex}
    \caption{Throughput in kilo-events per second (optical-flow type~\cite{Stoffregen17acra}) of Algorithm \ref{alg:ems} running on a single CPU core vs GPU for varying $\numLayers$ (using the test sequence in Fig.~\ref{fig:ijrr:depth}).
    }
    \label{tab:CPU_vs_GPU}
    \vspace{-1ex}
\end{table}
Regardless of throughput, our method allows exploiting key features of event cameras, such as its very high temporal resolution and its HDR, as shown in experiments on the EED dataset (Table~\ref{tab:iros18}) and on some sequences in Fig.~\ref{fig:allseqs} (Vehicle facing the sun, Fan and coin).

\subsection{Further Real-World Sequences}
We test our method on additional sequences in a variety of real-world scenes, as shown in Fig.~\ref{fig:allseqs}.
The third column shows the results of segmenting a traffic scene, with the camera placed parallel to a street and tilting upwards while vehicles pass by in front of it.
The algorithm segmented the scene into four clusters: three corresponding to the vehicles (blue, red, yellow) and another one corresponding to the background buildings (ego-motion, in green).
Even the cars going in the same direction are separated (red and yellow), since they travel at different speeds.

The fourth column of Fig.~\ref{fig:allseqs} shows the results of segmenting a scene with a car and some buildings while the camera is panning. 
We ran the algorithm to segment the scene into three clusters using optical flow warps. 
One cluster segments the car, and the other two clusters are assigned to the buildings.
Note that the algorithm detects the differences in optical flow due to the perspective projection of the panning camera: it separates the higher speed in the periphery (blue) from the slower speed in the image center (red).

An HDR scene is shown on the fifth column of Fig.~\ref{fig:allseqs}.
The camera is mounted on a moving vehicle facing the sun (central in field of view) while a pedestrian and a skateboarder cross in front of it.
The camera's forward motion causes fewer events from the background than in previous (panning) examples. 
We run the segmentation algorithm with six clusters, allowing the method to adapt to the scene. 
Segmented events are colored according to cluster membership.
The algorithm correctly segments the pedestrian and the skateboarder, 
producing motion-compensated images of their silhouettes despite being non-rigidly moving objects.

Finally, the last column of Fig.~\ref{fig:allseqs} shows the versatility of our method to accommodate different motion models for each cluster.
To this end, we recorded a coin dropping in front of the blades of a ventilator spinning at \SI{1800}{\revpermin}.
In this case the fan is represented by a rotational motion model and the coin by a linear velocity motion model. 
Our method converges to the expected, optimal solution, as can be seen in the motion compensated images, and it can handle the occlusions on the blades caused by the coin.

\global\long\def\highreswidth{0.42\linewidth}
\begin{figure}[t]
\centering
\subfloat[Scene]{\frame{\includegraphics[width=\highreswidth]{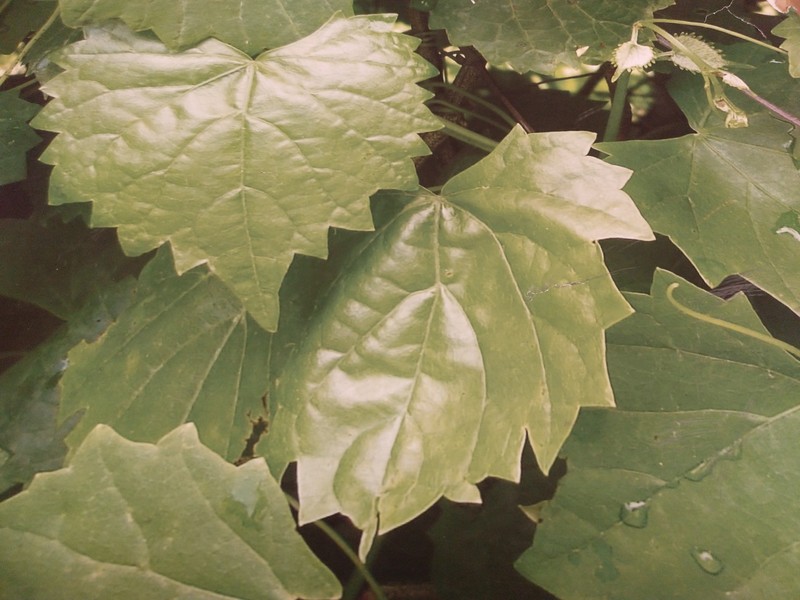}}}~~~
\subfloat[Segmentation into clusters]{\frame{\includegraphics[width=\highreswidth]{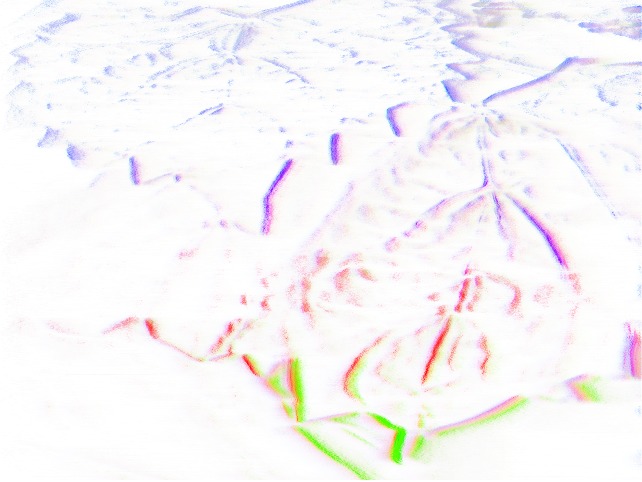}}}
\vspace{-1ex}
    \caption{\label{fig:high-res-solo} 
    A 640$\!\times\!$480 pixel DVS~\cite{Son17isscc} panning over a slanted plane. 
    Segmentation with 10 optical flow clusters (colored). 
    }
    \vspace{-2ex}
\end{figure}
Fig.~\ref{fig:high-res-solo} shows that our method also works with a higher resolution ($640\times480$ pixels) event-based camera~\cite{Son17isscc}.
More experiments are provided in the Appendix.

\subsection{Sensitivity to the Number of Clusters}
\label{sec:exp:num_clusters}
The following experiment shows that our method is not sensitive to the number of clusters chosen $\numLayers$.
We found that $\numLayers$ is not a particularly important parameter; if it chosen to be too large, the unnecessary clusters end up not having any events allocated to them and thus ``die''. 
This is a nice feature, since it means that in practice $\numLayers$ can simply be chosen to be large and then not be worried about.
We demonstrate this on the {\small\texttt{slider\_depth}} sequence from~\cite{Mueggler17ijrr}; where there are multiple objects at different depths (depth continuum), with the camera sliding past them.
Because of parallax, this results in a continuum of image plane velocities and thus infinitely many clusters would in theory be needed to segment the scene with an optical flow motion-model.
Thus the sequence can only be resolved by adding many clusters which discretize the continuum of velocities.

\global\long\def\sliderdepthwidth{0.42\linewidth}
\begin{figure}[t]
    \centering%
    \subfloat[Scene]{\frame{\includegraphics[width=\sliderdepthwidth]{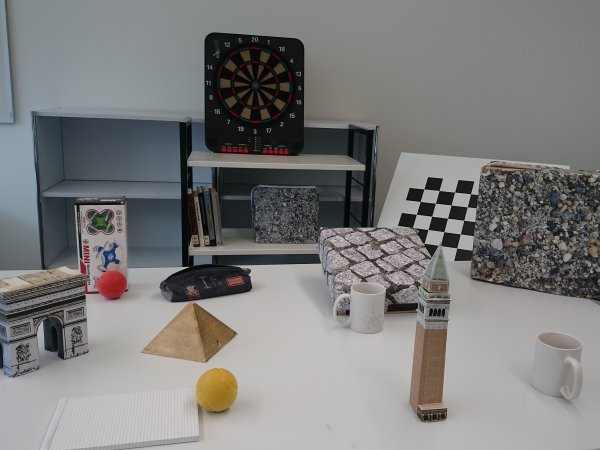}}}~~~~
    \subfloat[$\numLayers=5\times$Optical Flow]{%
    \frame{\includegraphics[width=\sliderdepthwidth]{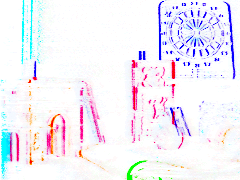}}
    \label{fig:ijrr:n5}}\\
    \subfloat[$\numLayers=10\times$Optical Flow]{%
    \frame{\includegraphics[width=\sliderdepthwidth]{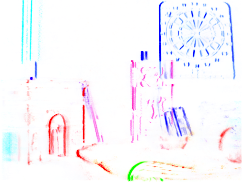}}
    \label{fig:ijrr:n15}}~~~
    \subfloat[$\numLayers=20\times$Optical Flow]{%
    \frame{\includegraphics[width=\sliderdepthwidth]{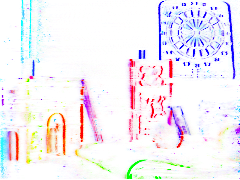}}
    \label{fig:ijrr:n20}}\\
    \subfloat[$\numLayers=5\times \text{OF} + 5\times \text{Rotation}$]{%
    \frame{\includegraphics[width=\sliderdepthwidth]{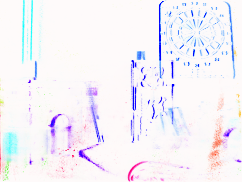}}
    \label{fig:ijrr:n505r}}~~
    \subfloat[$\numLayers=10\times$Rotation]{
    \frame{\includegraphics[width=\sliderdepthwidth]{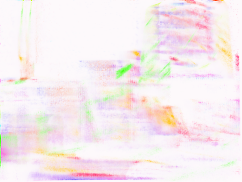}}
    \label{fig:ijrr:n10r}}
    \vspace{-1ex}
    \caption{\label{fig:ijrr:depth}
    \emph{Experiment on {\small\texttt{slider\_depth}} sequence} of~\cite{Mueggler17ijrr}.
    Motion-compensated images in \ref{fig:ijrr:n5} to \ref{fig:ijrr:n10r} show events colored by cluster.
    Using optical flow (OF) warps (in \ref{fig:ijrr:n5}--\ref{fig:ijrr:n20}), event clusters correspond to depth planes with respect to the camera.
    Using as few as five clusters, the events are discretized and approximately spread over the depth continuum.
    Using 5, 10 or 20 clusters in \ref{fig:ijrr:n5}, \ref{fig:ijrr:n15}, \ref{fig:ijrr:n20} gives very similar results, showing that our method is fairly insensitive to the value of $\numLayers$ chosen. 
    Adding clusters with motion models that do not suit the motion, as in \ref{fig:ijrr:n505r}, where five clusters are pure rotational warps, does not much disturb the output either.}
    \vspace{-3ex}
\end{figure}
Fig. \ref{fig:ijrr:depth} demonstrates that our method is robust with regard to the number of clusters chosen (in Figs.~\ref{fig:ijrr:n5}--\ref{fig:ijrr:n20}); 
too few clusters and the method will simply discretize the event cluster continuum, too many clusters and some clusters will ``collapse'', i.e., no events will be assigned to them.
By segmenting with enough clusters and preventing cluster collapse, our method can be used to detect depth variations;
nevertheless, tailored methods for depth estimation~\cite{Rebecq17ijcv} are more suitable for such a task.
The experiment also shows that our method deals with object occlusions.

Similarly, Fig.~\ref{fig:ijrr:depth} shows that our method is not sensitive to the mixture of motion models either.
Fig.~\ref{fig:ijrr:n505r} shows the result with five clusters of optical flow type and five clusters of rotation type. 
As can be seen, our method essentially allocates no event likelihoods to these rotation models clusters, which clearly do not suit any of the events in this sequence.
Fig.~\ref{fig:ijrr:n10r} shows the result of using only rotation motion models, resulting in failure, as expected.
As future work, a meta-algorithm could be used to select which motion models are most relevant depending on the scene.

\vspace{-1ex}
\section{Conclusion}
\label{sec:conclusion}
\vspace{-1ex}
In this work we presented the first method for per-event segmentation of a scene into multiple objects based on their apparent motion on the image plane.
We jointly segmented a given set of events and recovered the motion parameters of the different objects (clusters) causing them. 
Additionally, as a by-product, our method produced motion-compensated images with the sharp edge-like appearance of the objects in the scene, which may be used for further analysis (e.g., recognition).
We showed that our method outperforms two recent methods on a publicly available dataset (with as much as \SI{10}{\percent} improvement over the state-of-the-art~\cite{Mitrokhin18iros}),
and showed it can resolve small relative motion differences between clusters.
Our method achieves this using a versatile cluster model and avoiding explicit estimation of optical flow for motion segmentation, which is error prone.
All this allowed us to perform motion segmentation on challenging scenes, such as high speed and/or HDR, unlocking the outstanding properties of event-based cameras.
\vspace{-3.5ex}
\paragraph*{Acknowledgments.} This work was supported by the Swiss National Center of Competence in Research Robotics, through the Swiss National Science Foundation and the SNSF-ERC Starting Grant as well as the ARC Centre of Excellence for Robot Vision, project \#CE140100016.

\appendix
\cleardoublepage
\title{\MYTITLE\\---Supplementary Material---}
\maketitle

\section{Multimedia Material}
\label{sec:supplementary}
The video accompanying this work is available at\\
\url{https://youtu.be/0q6ap_OSBAk}

\section{Two Additional Motion-Compensation Segmentation Methods}

In this section, we describe how two classical clustering methods (mixture densities and fuzzy k-means) can be modified to tackle the task of event-based motion segmentation, by leveraging the idea of motion-compensation~\cite{Gallego17ral,Gallego18cvpr} (Sections~\ref{sec:MixtureDensitiesEvent} and~\ref{sec:FuzzykMeansEvent}).
Examples comparing the three per-event segmentation models developed (Algorithms~\ref{alg:ems} to~\ref{alg:fuzzyk}) are given in Section~\ref{sec:threemodel:comparison}; 
they are called \emph{proposed} (or \emph{Layered}), \emph{Mixture Densities} and \emph{Fuzzy k-Means}, respectively.

\subsection{Mixture Densities}
\label{sec:MixtureDensitiesEvent}
The mixture models framework~\cite{Duda00book,Bishop06book} can be adapted to solve the segmentation problem addressed.
The idea is to fit a mixture density to the events $\cE$, with each mode representing a cluster of events with a coherent motion.

\paragraph{Problem Formulation.}
Specifically, following the notation in~\cite[Ch.10]{Duda00book}, we identify the elements of the problem:
the data points are the events $\cE$ without taking into account polarity; 
thus, feature space is the volume $V$, and, consequently, the clusters are comprised of events (i.e., they are not clusters of optic flow vectors in velocity space).

The mixture model states that events $e_{k}\in V$ are distributed according to a sum of several distributions (``clusters''), 
with mixing weights (``cluster probabilities'') $\bpi \doteq \{P(\omega_{j})\}_{j=1}^{\numLayers}$:
\begin{equation}
p(e_{k} | \bparams) = \sum_{j=1}^{\numLayers} p(e_{k} | \omega_{j},\bparams_{j}) P(\omega_{j}),
\label{eq:mixtudeDensity}
\end{equation}
where $\bparams = \{\bparams_{j}\}_{j=1}^{\numLayers}$ are the parameters of the distributions of each component of the mixture model and 
we assumed that the parameters of each cluster are independent of each other: 
$p(e_{k} | \omega_{j},\bparams) = p(e_{k} | \omega_{j},\bparams_{j})$.
The function $p(z | \bparams)$ in~\eqref{eq:mixtudeDensity}, with $z\in V$, is a scalar field in $V$ representing the density of events in $V$ as a sum of several densities, each of them corresponding to a different cluster, and each cluster describing a coherent motion.

To measure how well the $j$-th cluster explains an event~\eqref{eq:mixtudeDensity}, 
we propose to use the unweighted IWE~\eqref{eq:unweightedIWE}: 
\begin{eqnarray}
\label{eq:singleEventLikelihood}
p(z \,| \, \omega_{j},\bparams_{j}) & \propto & H_j \left(\bx'(z;\bparams_{j})\right)\\
\label{eq:unweightedIWE}
H_j \left(\bx\right) & \doteq & \sum_{m=1}^{\numEvents} \delta \left(\bx - \bx'_{mj}\right)
\end{eqnarray}
with warped event location $\bx'_{mj} = \Warp(\bx_{m},t_{m};\bparams_j)$.
The image point $\bx'(z;\bparams_{j})$ corresponds to the warped location of point $z\in V$ using the motion parameters of the $j$-th cluster. 

Hence, the goodness of fit between a point $z\in V$ and the $j$-th cluster is measured by the amount of event alignment (i.e., ``sharpness''): the larger the IWE of the cluster at the warped point location, the larger the probability that $z$ belongs to the cluster.

Notice that the choice~\eqref{eq:singleEventLikelihood} makes the distribution of each component in the mixture $p(z \,| \, \omega_{j},\bparams_{j})$ be constant along the point trajectories defined by the warping model of the cluster, 
which agrees with the ``tubular'' shape mentioned in the problem statement (Section~\ref{sec:method}).
The mixture model~\eqref{eq:mixtudeDensity} may not be constant along point trajectories since it is a weighted sum of several distributions, each with its own point trajectories.

\paragraph{Iterative Solver: EM Algorithm.}
With the above definitions, we may apply the EM algorithm in~\cite[Ch.10]{Duda00book} to compute the parameters of the mixture model, 
by maximizing the (log-)likelihood of the mixture density:
\begin{equation}
\label{eq:logLikelihoodArgMax}
\left(\bparams^\ast,\bpi^\ast\right) = \argmax_{(\bparams,\bpi)} \sum_{k=1}^{\numEvents}\log\, p(e_{k} | \bparams)
\end{equation}

\begin{algorithm}[t!]
\caption{Event-based Motion Segmentation using Mixture Density Model}
\label{alg:mixmodel}
\begin{algorithmic}[1]
\State \textbf{Input}: events $\cE=\{e_k\}_{k=1}^{\numEvents}$ in a space-time volume $V$ of the image plane, and number of clusters $\numLayers$.
\State \textbf{Output}: cluster parameters $\bparams=\{\bparams_j\}_{j=1}^{\numLayers}$ 
and mixing weights $\bpi \doteq \{P(\omega_{j})\}_{j=1}^{\numLayers}$.
\State \textbf{Procedure:}
\State Initialize $\bparams$ and $\bpi$.
\State \textbf{Iterate} until convergence:
  \State $\bullet$ Update the mixing weights~\eqref{eq:MixModelUpdateMixingWeights}, using the current motion parameters $\bparams$ and the mixing weights from the previous iteration in \eqref{eq:MixModelMembershipProb}. 
  \State $\bullet$ Update motion parameters $\bparams$ by ascending on~\eqref{eq:logLikelihoodArgMax}.
\end{algorithmic}
\end{algorithm}

In the E-step, the mixing weights $\bpi$ are updated using
\begin{equation}
\label{eq:MixModelUpdateMixingWeights}
P(\omega_{j}) = \frac{1}{\numEvents}\sum_{k=1}^{\numEvents} p(\omega_{j} | e_{k},\bparams_{j})
\end{equation}
with membership probabilities given by the Bayes formula
\begin{equation}
\label{eq:MixModelMembershipProb}
p(\omega_{j}|e_{k},\bparams_{j}) = \frac{p(e_{k}|\omega_{j},\bparams_{j})P(\omega_{j})} {\sum_{i=1}^{\numLayers}p(e_{k}|\omega_{i},\bparams_{i})P(\omega_{i})}.
\end{equation}
In the M-step, gradient ascent or conjugate gradient~\cite{Nocedal06book} of the log-likelihood~\eqref{eq:logLikelihoodArgMax} with respect to the warp parameters $\bparams$ is used to update $\bparams$, in preparation for the next iteration.

The pseudo-code of this mixture model method is given in Algorithm~\ref{alg:mixmodel}.
From the mixing weights and the motion parameters, it is straightforward to compute the event-cluster assignment probabilities 
using~\eqref{eq:MixModelMembershipProb}.
To initialize the iteration, we use the procedure described in Section~\ref{sec:method:initialization}.

Notice that, during the EM iterations, the above method not only estimates the cluster parameters $\bparams$ and the mixing weights $\bpi$ but also the distributions $p(z | \omega_{i},\bparams_{i})$ themselves, i.e., the ``shape'' of the components of the mixture model. 
These distributions get sharper (more peaky or ``in focus'') around the segmented objects as iterations proceed, and blurred around the non-segmented objects corresponding to that cluster. 
An example is given in Section~\ref{sec:threemodel:comparison}.

\subsection{Fuzzy k-Means}
\label{sec:FuzzykMeansEvent}

Event-based motion segmentation can also be achieved by designing an objective function similar to the one used in the fuzzy k-means algorithm~\cite[Ch.10]{Duda00book}.

\paragraph{Problem Formulation.}
This approach seeks to maximize
\begin{equation}
\label{eq:ObjFuncFuzzyKMeans}
(\bparams^\ast, \probmat^\ast) = \argmax_{\bparams,\probmat} \sum_{j=1}^{\numLayers} \sum_{k=1}^{\numEvents} p_{kj}^{b} d_{kj},
\end{equation}
where $b>1$ (e.g., $b=2$) adjusts the blending of the different clusters, 
and the goodness of fit between an event $e_k$ and a cluster $j$ in $V$ is given in terms of event alignment (i.e., ``sharpness''): 
\begin{equation}
\label{eq:fuzzy:distance_kj}
d_{kj} \doteq \log H_j (\bx'_{kj}),
\end{equation}
the value of the unweighted IWE~\eqref{eq:unweightedIWE} at the warped event location using the motion parameters of the cluster.
We use the logarithm of the IWE, as in~\eqref{eq:logLikelihoodArgMax}, to decrease the influence of large values of the IWE, since these are counted multiple times if the events are warped to the same pixel location.
Notice that~\eqref{eq:ObjFuncFuzzyKMeans} differs from~\eqref{eq:weightedIWE}-\eqref{eq:weightedContrast}: 
the responsibilities $p_{kj}$ appear multiplying the IWE (i.e., they are not included in a weighted IWE), 
and the sum is over the events (as opposed to over the pixels~\eqref{eq:ContrastOneLayer}).

Notice also that this proposal is different from clustering in optical flow space (Fig.~\ref{fig:rocks6:flowclusters}). 
As mentioned in Section~\ref{sec:MixtureDensitiesEvent}, here the feature space is the space-time volume $V\in\R^3$ (i.e., event location), rather than the optical flow space ($\R^2$) (i.e., event velocity).

\paragraph{Iterative Solver: EM Algorithm.}
The EM algorithm may also be used to solve~\eqref{eq:ObjFuncFuzzyKMeans}.
In the E-step (fixed warp parameters $\bparams$) the responsibilities are updated using the closed-form partitioning formula
\begin{equation}
\label{eq:FuzzyKmeansDistanceEstep}
p_{kj} = d_{kj}^{\frac{1}{b-1}} \left/ \sum_{i=1}^{\numLayers} d_{ki}^{\frac{1}{b-1}}\right. .
\end{equation}
In the M-step (fixed responsibilities) the warp parameters of the clusters $\bparams$ are updated using gradient ascent or conjugate gradient.
The pseudo-code of the event-based fuzzy k-means segmentation method is given in Algorithm~\ref{alg:fuzzyk}.
\begin{algorithm}[t!]
\caption{Event-based Motion Segmentation using the Fuzzy k-Means Method}
\label{alg:fuzzyk}
\begin{algorithmic}[1]
\State \textbf{Input}: events $\cE=\{e_k\}_{k=1}^{\numEvents}$ in a space-time volume $V$ of the image plane, and number of clusters $\numLayers$.
\State \textbf{Output}: cluster parameters $\bparams=\{\bparams_j\}_{j=1}^{\numLayers}$ and event-cluster assignments $\probmat \equiv p_{kj} \doteq P(e_k \in \layer_j)$.
\State \textbf{Procedure:}
\State Initialization (as in Section~\ref{sec:method:initialization}).
\State \textbf{Iterate} until convergence:
  \State $\bullet$ Update the event-cluster assignments $p_{kj}$ using~\eqref{eq:FuzzyKmeansDistanceEstep}.
    \State $\bullet$ Update motion parameters $\bparams$ by ascending on~\eqref{eq:ObjFuncFuzzyKMeans}.
\end{algorithmic}
\end{algorithm}

\subsection{Comparison of Three Motion-Compensation Segmentation Methods}
\label{sec:threemodel:comparison}
We compare our method with the two above-mentioned methods (Sections~\ref{sec:MixtureDensitiesEvent} and \ref{sec:FuzzykMeansEvent}) that we also designed to leverage motion compensation.

\global\long\def\layerwidth{0.29\linewidth}

\begin{figure}[t]
	\centering
    {\small
    \setlength{\tabcolsep}{2pt}
	\begin{tabular}{  >{\centering\arraybackslash}m{0.6cm} 
	>{\centering\arraybackslash}m{\layerwidth} 
	>{\centering\arraybackslash}m{\layerwidth} 
	>{\centering\arraybackslash}m{\layerwidth}}
	    & Cluster 1 & Cluster 2 & Cluster 3\\
		\rotatebox{90}{\makecell{
		Proposed\\[-0.5ex]
		method}}	    
		&
		\frame{\includegraphics[width=\linewidth]{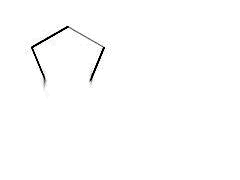}}
		&
		\frame{\includegraphics[width=\linewidth]{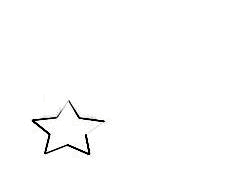}}
		&
		\frame{\includegraphics[width=\linewidth]{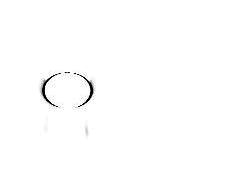}}
		\\
		\rotatebox{90}{\makecell{
		Mixture\\[-0.5ex]
		density}}
		&
		\frame{\includegraphics[width=\linewidth]{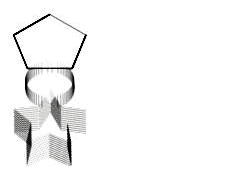}}
		&
		\frame{\includegraphics[width=\linewidth]{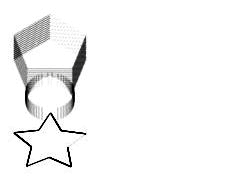}}
		&
		\frame{\includegraphics[width=\linewidth]{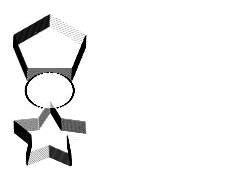}}
		\\
		\rotatebox{90}{\makecell{
		Fuzzy\\[-0.5ex]
		k-means}}
		&
		\frame{\includegraphics[width=\linewidth]{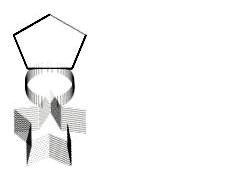}}
		&
		\frame{\includegraphics[width=\linewidth]{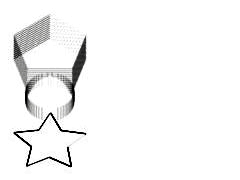}}
		&
		\frame{\includegraphics[width=\linewidth]{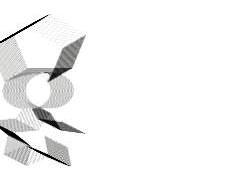}}\\
	\end{tabular}
	}
	\caption{Comparison of three methods for event-based Motion Segmentation: Algorithms~\ref{alg:ems} to~\ref{alg:fuzzyk} (one per row).
	}
	\label{fig:comparisonSegmentationMethods}
\end{figure}
Fig.~\ref{fig:comparisonSegmentationMethods} shows the comparison of the three methods on a toy example with three objects (a filled pentagon, a star and a circle) moving in different directions on the image plane.
In the mixture density and fuzzy k-means methods, the motion-compensated IWEs do not include the event-cluster associations $\probmat$, and so, all objects appear in all IWEs, sharper in one IWE than in the others.
In contrast, in the proposed method (Algorithm~\ref{alg:ems}), the associations are included in the motion-compensated image of the cluster (weighted IWE), as per equation~\eqref{eq:weightedIWE}, and so, the objects are better split into the clusters (with minor ``ghost'' effects, as illustrated in Fig.~\ref{fig:method:layers}), thus yielding the best results.

It is worth mentioning that the three per-event segmentation methods are novel: they have not been previously proposed in the literature. 
We decided to focus on Algorithm~\ref{alg:ems} in the main part of the paper and thus leave the adaptation of classical methods (mixture model and fuzzy k-means) for the supplementary material.

\subsection{Computational Complexity}
Next, we analyze the complexity of the three segmentation methods considered (Algorithms~\ref{alg:ems} to~\ref{alg:fuzzyk}), 
defined by objective functions~\eqref{eq:weightedContrast}, \eqref{eq:logLikelihoodArgMax} and \eqref{eq:ObjFuncFuzzyKMeans}, respectively.
The core of the segmentation methods is the computation of the images of warped events (IWEs \eqref{eq:weightedIWE} or \eqref{eq:unweightedIWE}; one per cluster), which has complexity $O(\numEvents \numLayers)$.

\paragraph{Proposed (Layered) Model.}
The complexity of updating the event assignments using~\eqref{eq:probEventInLayer} is essentially that of computing the (weighted) IWEs of all clusters, i.e., $O(\numEvents \numLayers)$.
The complexity of computing the contrast~\eqref{eq:ContrastOneLayer} of a generic image is linear in the number of pixels, $O(\numPixels)$, and so, the complexity of computing the contrast of one IWE is $O(\numEvents + \numPixels)$.
The computation of the contrast is negligible compared to the effort required by the warp.
Computing the contrast of $\numLayers$ clusters (corresponding to a set of candidate parameters) has complexity $O\bigl( (\numEvents + \numPixels)\numLayers \bigr)$.
Since multiple iterations $\numIters$ may be required to find the optimal parameters, the total complexity of the iterative algorithm used is $O\bigl( (\numEvents + \numPixels)\numLayers \numIters \bigr)$.

\paragraph{Mixture Density Model.}
The complexity of updating the mixture weights is that of computing the posterior probabilities $p(\omega_{j}|e_{k},\bparams_{j})$, which require to compute the IWEs of all clusters, i.e., complexity $O(\numEvents\numLayers)$.
The complexity of updating the motion parameters is also that of computing the contrasts of the IWEs of all clusters, through multiple ascent iterations. 
In total, the complexity is $O\bigl((\numEvents+\numPixels)\numLayers\numIters\bigr)$.

\paragraph{Fuzzy k-means Model.}
The complexity of computing the responsibilities~\eqref{eq:FuzzyKmeansDistanceEstep} is that of computing $\numLayers$ IWEs (values $d_{kj}$), i.e., $O(\numEvents\numLayers)$.
The complexity of updating the motion parameters is that of computing the objective function~\eqref{eq:ObjFuncFuzzyKMeans}, $O(\numEvents\numLayers)$, through multiple iterations. 
In total, the complexity is $O(\numEvents\numLayers\numIters)$.


\begin{figure}[t]
    \centering
    \subfloat[Value of the sum-of-contrasts objective function~\eqref{eq:weightedContrast} for each of the three  methods per iteration of optimization.\label{fig:comp:ours}]{\includegraphics[width=0.95\linewidth]{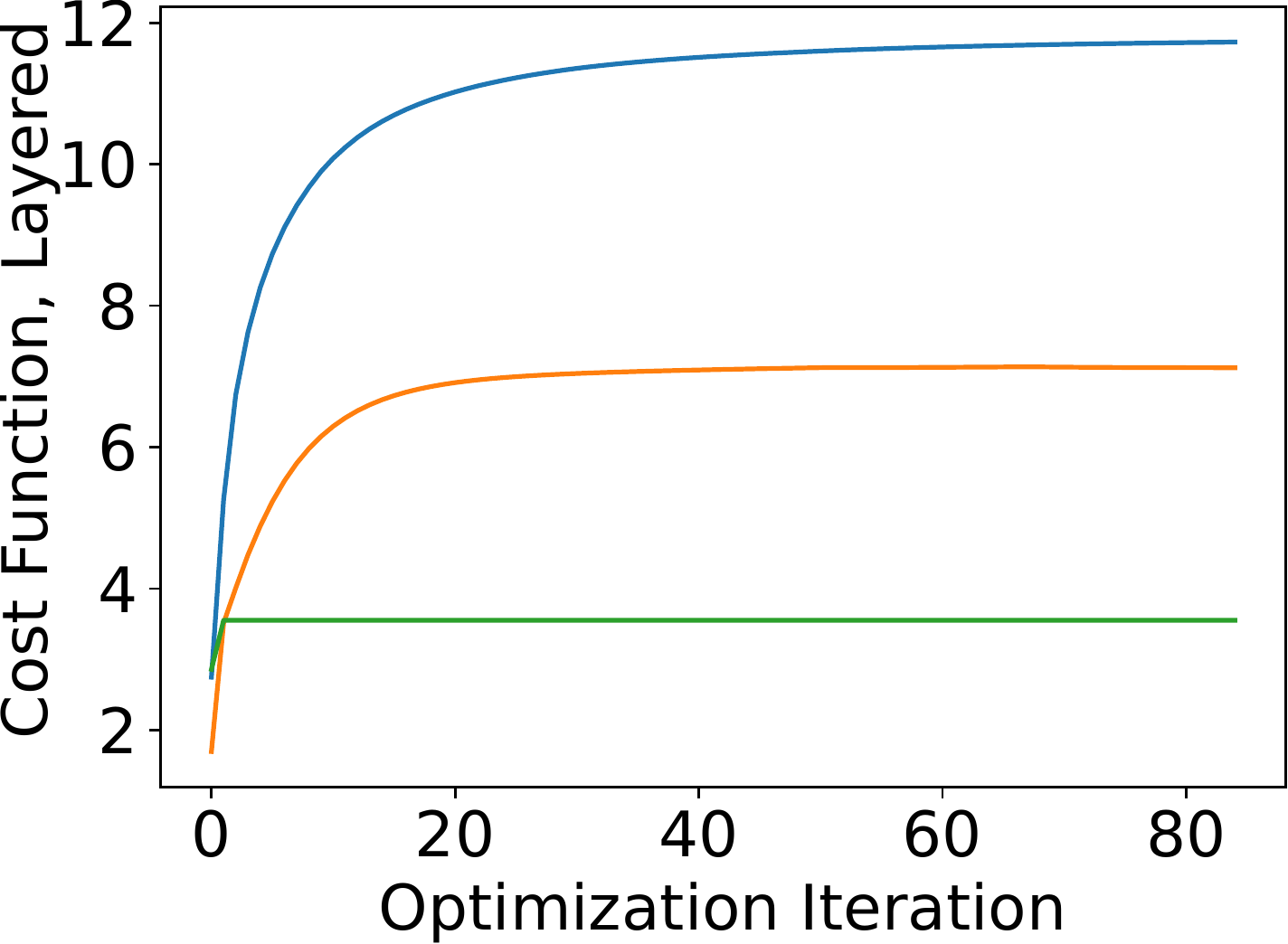}}\\
    \subfloat[Number of warps performed by each method per iteration of optimization.\label{fig:comp:warp_iter}]{\includegraphics[width=0.95\linewidth]{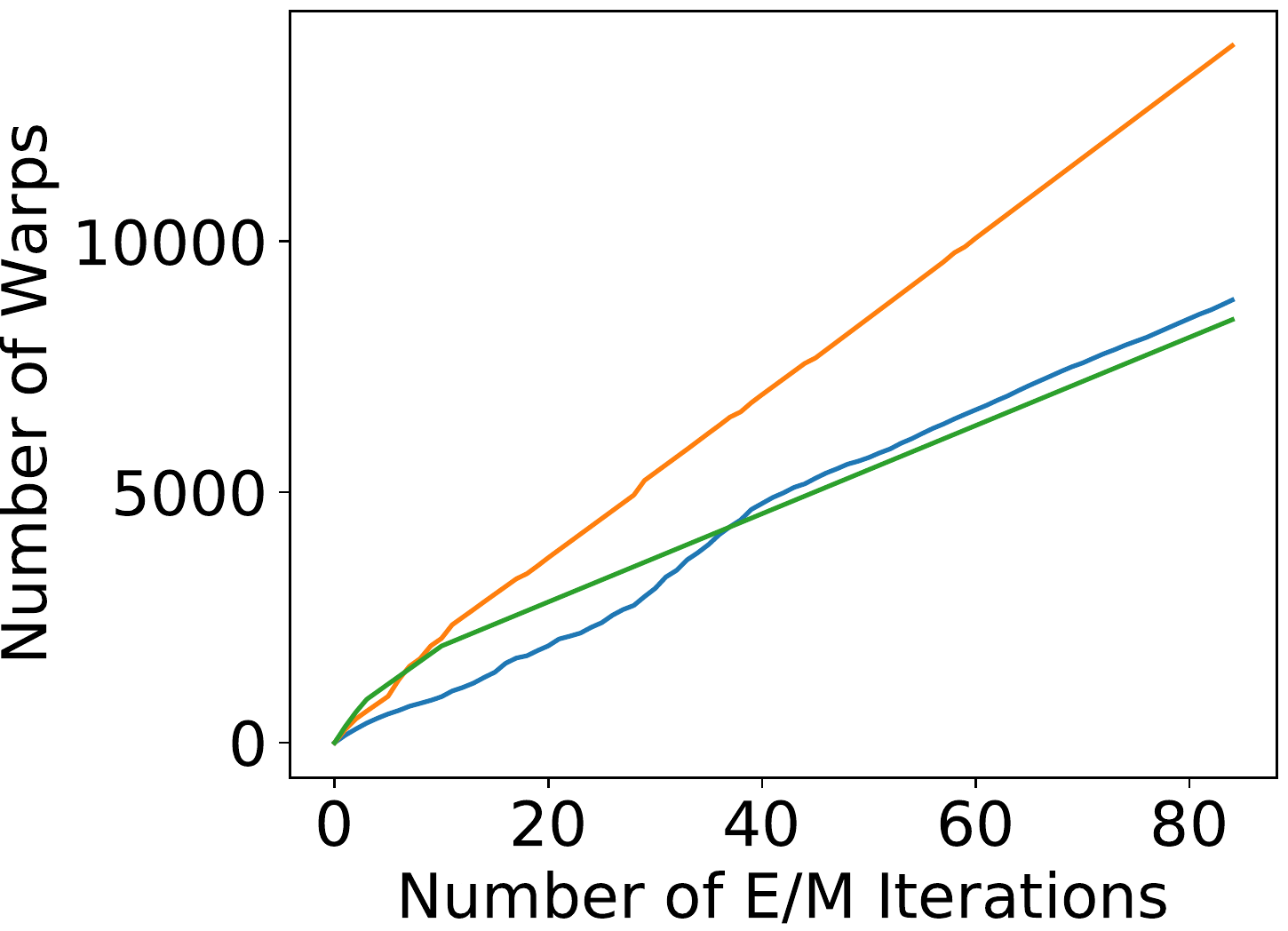}}
    \caption{\label{fig:comparison}
    \emph{Comparison of three Methods.} 
    We compare the convergence properties of three motion-compensated event-segmentation methods (Algorithms~\ref{alg:ems} to~\ref{alg:fuzzyk}): proposed (layered) method (\emph{blue}), Mixture Density Model (\emph{orange}) and Fuzzy k-Means (\emph{green}).
    Data used is from the Traffic Sequence (third column of Fig. \ref{fig:allseqs}), the warped events at each iteration are visualized in Fig. \ref{fig:ours_mm_fkm_convergence}.}
\end{figure}

\global\long\def\carsconvergwidth{0.191\linewidth}
\begin{figure*}[t]
\centering
\subfloat[Proposed, $\text{iter} = 1$]{\frame{\includegraphics[width=\carsconvergwidth]{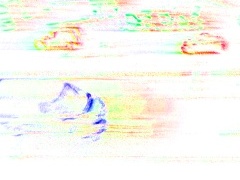}}}~
\subfloat[$\text{iter} = 5$]{\frame{\includegraphics[width=\carsconvergwidth]{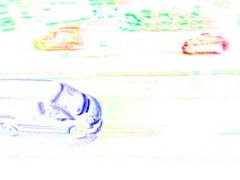}}}~
\subfloat[$\text{iter} = 10$]{\frame{\includegraphics[width=\carsconvergwidth]{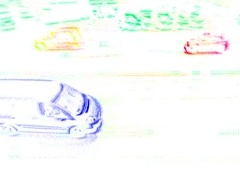}}}~
\subfloat[$\text{iter} = 20$]{\frame{\includegraphics[width=\carsconvergwidth]{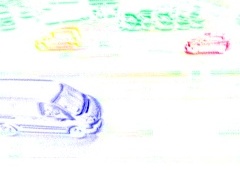}}}~
\subfloat[$\text{iter} = 80$]{\frame{\includegraphics[width=\carsconvergwidth]{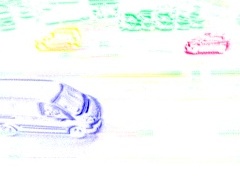}}}\\
\subfloat[Mixture Density, $\text{iter} = 1$]{\frame{\includegraphics[width=\carsconvergwidth]{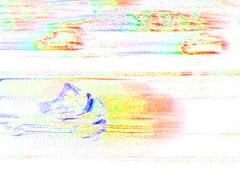}}}~
\subfloat[$\text{iter} = 5$]{\frame{\includegraphics[width=\carsconvergwidth]{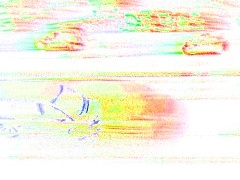}}}~
\subfloat[$\text{iter} = 10$]{\frame{\includegraphics[width=\carsconvergwidth]{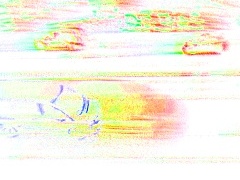}}}~
\subfloat[$\text{iter} = 20$]{\frame{\includegraphics[width=\carsconvergwidth]{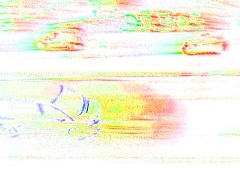}}}~
\subfloat[$\text{iter} = 80$]{\frame{\includegraphics[width=\carsconvergwidth]{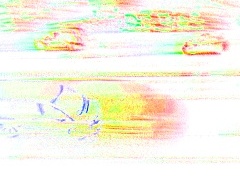}}}\\
\subfloat[Fuzzy k-Means, $\text{iter} = 1$]{\frame{\includegraphics[width=\carsconvergwidth]{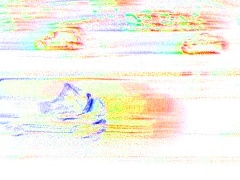}}}~
\subfloat[$\text{iter} = 5$]{\frame{\includegraphics[width=\carsconvergwidth]{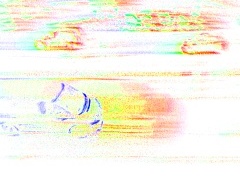}}}~
\subfloat[$\text{iter} = 10$]{\frame{\includegraphics[width=\carsconvergwidth]{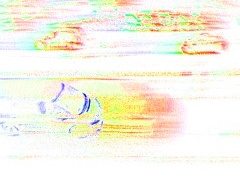}}}~
\subfloat[$\text{iter} = 20$]{\frame{\includegraphics[width=\carsconvergwidth]{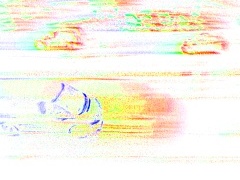}}}~
\subfloat[$\text{iter} = 80$]{\frame{\includegraphics[width=\carsconvergwidth]{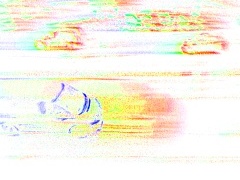}}}    
    \caption{\label{fig:ours_mm_fkm_convergence}Images of the motion-corrected events for three segmentation methods. 
    From left to right the images show the state after 1, 5, 10, 20 and 80 iterations respectively.
    \emph{Top Row}: Algorithm~\ref{alg:ems},
    \emph{Middle Row}: Algorithm~\ref{alg:mixmodel},
    \emph{Bottom Row}: Algorithm~\ref{alg:fuzzyk}.}
\end{figure*}

\paragraph{Plots of Computational Effort and Convergence.}
Fig.~\ref{fig:comparison} shows the convergence of the three above methods on real data from a traffic sequence that is segmented into four clusters (Fig. \ref{fig:ours_mm_fkm_convergence} and third column of Fig.~\ref{fig:allseqs}): three cars and the background due to ego-motion.
The top plot, Fig.~\ref{fig:comp:ours}, shows the evolution of the sum-of-contrasts objective function~\eqref{eq:weightedContrast} vs the iterations. 
All methods stagnate after $\approx 20$ iterations, and, as expected, the proposed method provides the highest score among all three methods (since it is designed to maximize this objective function).
The Mixture model and Fuzzy k-means methods do not provide such a large score mostly due to the event-cluster associations, since they are not as confident to belonging to one cluster as in the proposed method.
Fig.~\ref{fig:comp:warp_iter} displays the number of warps (i.e., number of IWEs) that each method computes as the optimization iterations proceed; 
as it can be shown, the relationship is approximately linear, 
with the proposed method performing the least warps for a considerable number of iterations, before stagnation (Fig.~\ref{fig:comp:ours}).

\section{Additional Experiments}
\label{sec:suppl_experiments}

\subsection{Non-rigid Moving Objects}
\label{subsec:exp:non-rigid}
In the following experiments we show how our method deals with non-rigid objects.
Our algorithm warps events $\cE$ according to point-trajectories described by parametric motion models whose parameters are assumed constant over the (small) time $\Delta t$ spanned by $\cE$.
Low-dimensional parametric warp models, such as the patch-based optic flow (2-DOF, linear trajectories), rotational motion in the plane (1-DOF) or in space (3-DOF) are simple and 
produce robust results by constraining the dimensionality of the solution space in the search for optimal point-trajectories. 
However, simple warp models (both in event-based vision or in traditional frame-to-frame vision) have limited expressiveness: they are good for representing rigidly moving objects, but do not have enough degrees of freedom to represent more complex motions, such as deformations (e.g., pedestrian, birds, jelly fish, etc.).
One could consider using warps able to describe more complex motions, such as part-based warp models~\cite{Kumar08ijcv} or infinite-dimensional models~\cite{Yezzi03ijcv}. 
But this would make the segmentation problem considerably harder, not only due to the increased dimensionality of the search space, but also because it would be possibly filled with multiple local minima.

\paragraph*{Pedestrian.}
Fig.~\ref{fig:walkTimo} shows a pedestrian walking past the camera while it is panning.
In spite of using simple warp models, our method does a good job at segmentation: the background (due to camera motion), the torso of the person and the swinging arms are segmented in separate clusters.
This is so, because during the short time span of $\cE$ (in the order of milliseconds), the objects move approximately rigidly.
\global\long\def\walkingwidth{0.4\linewidth}
\begin{figure*}[t]
    \centering
    \subfloat[Cluster 1]{\frame{\includegraphics[width=\walkingwidth]{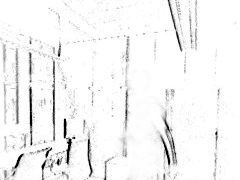}\label{fig:walkTimo:room}}}~~
    \subfloat[Cluster 2]{\frame{\includegraphics[width=\walkingwidth]{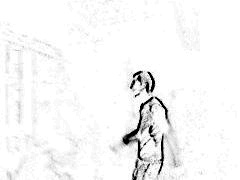}\label{fig:walkTimo:torso}}}\\[-1ex]
    \subfloat[Cluster 3]{\frame{\includegraphics[width=\walkingwidth]{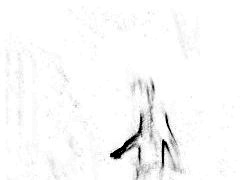}\label{fig:walkTimo:arms}}}~~
    \subfloat[All clusters (merged IWEs)]{\frame{\includegraphics[width=\walkingwidth]{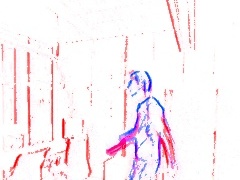}\label{fig:walkTimo:global}}}
    \caption{\label{fig:walkTimo}\emph{Non-Rigid Scene.}
    A person walks across a room, arms swinging. The room \ref{fig:walkTimo:room}, the body \ref{fig:walkTimo:torso} and the arms \ref{fig:walkTimo:arms} are segmented out, with greater uncertainty to the event associations in areas of deformation (such as elbows), visible in the fact that events are associated to both clusters (events colored by cluster in \ref{fig:walkTimo:global}).}
\end{figure*}

\paragraph*{Popping Balloon.}
In order to test the limits of this assumption, we filmed the popping of a balloon with the event camera (see Fig.~\ref{fig:balloon}).
While the segmentation struggles to give a clear result in the initial moments of puncturing (\ref{fig:balloon:b2}), it still manages to give reasonable results for the fast moving, contracting fragments of rubber flung away by the explosion (\ref{fig:balloon:b3}, \ref{fig:balloon:b4}).
\global\long\def\widthFourSnapshots{0.24\linewidth}
\begin{figure*}[t]
    \centering
    \subfloat[\label{fig:balloon:b1}Animated]{%
    \ifcompileanimated
    \frame{\animategraphics[width=\widthFourSnapshots]{15}{images/balloon/baloon_all_frames/global-}{01}{39}}
    \else
    \frame{\includegraphics[width=\widthFourSnapshots]{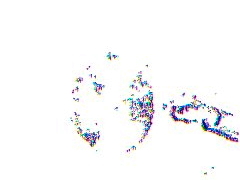}}
    \fi
    }~
    \subfloat[\label{fig:balloon:b2}]{\frame{\includegraphics[width=\widthFourSnapshots]{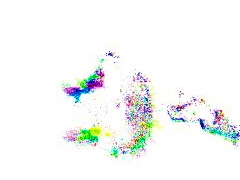}}}~
    \subfloat[\label{fig:balloon:b3}]{\frame{\includegraphics[width=\widthFourSnapshots]{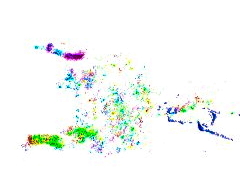}}}~
    \subfloat[\label{fig:balloon:b4}]{\frame{\includegraphics[width=\widthFourSnapshots]{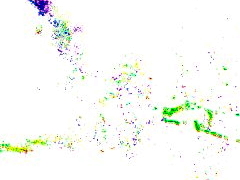}}}
    \caption{\label{fig:balloon}
    \emph{Non-Rigid Moving Objects.}
    From left to right: snapshots of segmentation of balloon popping. Run with $\numLayers=4$ clusters, events colored by cluster membership.}
\end{figure*}

\subsection{Additional to Section~\ref{sec:exp:num_clusters} - Continuum Depth Variation}
In this experiment we essentially show a scene similar to that in Fig.~\ref{fig:ijrr:depth}.
The difference is that the scene in Fig.~\ref{fig:checkers} shows a truly continuous depth variation.
As can be seen in the results (using $\numLayers = 15$), our method discretizes the segmentation, although it is noteworthy that each ``slice'' of depth appears to fade toward foreground and background.
This is because the method becomes less certain of the likelihood of events that sit between clusters, the darkness of a region reflecting the likelihood of a given event belonging to that cluster.
\global\long\def\checkerswidth{0.18\linewidth}
\begin{figure*}[t]
\centering
\subfloat[\label{fig:checkers_02}]{\frame{\includegraphics[width=\checkerswidth]{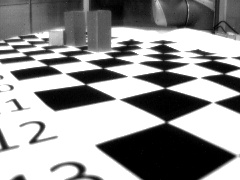}}}~
\subfloat[\label{fig:checkers_03}]{\frame{\includegraphics[width=\checkerswidth]{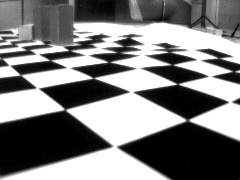}}}~
\subfloat[\label{fig:checkers_04}]{\frame{\includegraphics[width=\checkerswidth]{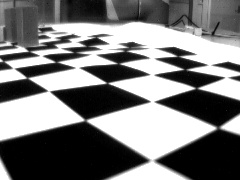}}}~
\subfloat[\label{fig:checkers_05}]{\frame{\includegraphics[width=\checkerswidth]{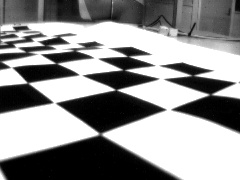}}}~
\subfloat[\label{fig:checkers_all}]{\frame{\includegraphics[width=\checkerswidth]{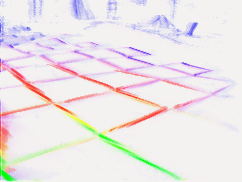}}}\\
\subfloat[\label{fig:checkers_06}]{\frame{\includegraphics[width=\checkerswidth]{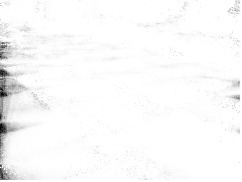}}}~
\subfloat[\label{fig:checkers_07}]{\frame{\includegraphics[width=\checkerswidth]{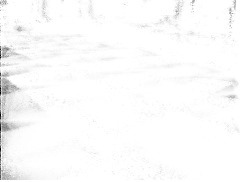}}}~
\subfloat[\label{fig:checkers_08}]{\frame{\includegraphics[width=\checkerswidth]{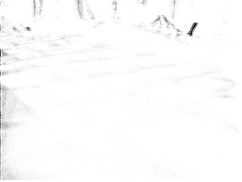}}}~
\subfloat[\label{fig:checkers_09}]{\frame{\includegraphics[width=\checkerswidth]{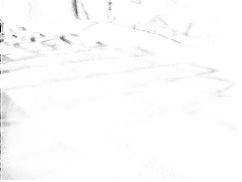}}}~
\subfloat[\label{fig:checkers_10}]{\frame{\includegraphics[width=\checkerswidth]{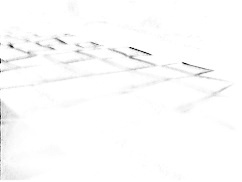}}}\\
\subfloat[\label{fig:checkers_11}]{\frame{\includegraphics[width=\checkerswidth]{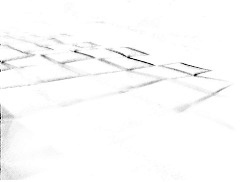}}}~
\subfloat[\label{fig:checkers_12}]{\frame{\includegraphics[width=\checkerswidth]{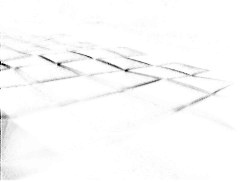}}}~
\subfloat[\label{fig:checkers_13}]{\frame{\includegraphics[width=\checkerswidth]{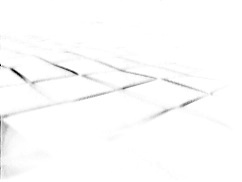}}}~
\subfloat[\label{fig:checkers_14}]{\frame{\includegraphics[width=\checkerswidth]{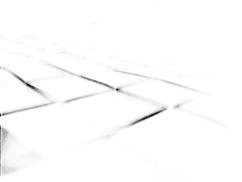}}}~
\subfloat[\label{fig:checkers_15}]{\frame{\includegraphics[width=\checkerswidth]{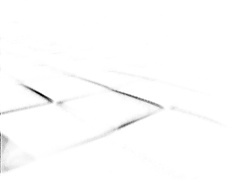}}}\\
\subfloat[\label{fig:checkers_16}]{\frame{\includegraphics[width=\checkerswidth]{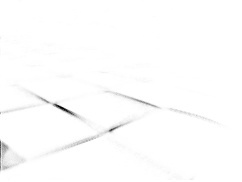}}}~
\subfloat[\label{fig:checkers_17}]{\frame{\includegraphics[width=\checkerswidth]{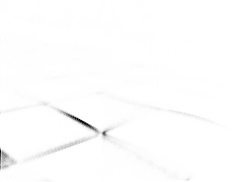}}}~
\subfloat[\label{fig:checkers_18}]{\frame{\includegraphics[width=\checkerswidth]{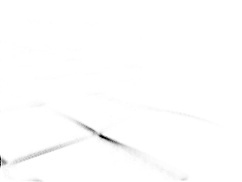}}}~
\subfloat[\label{fig:checkers_19}]{\frame{\includegraphics[width=\checkerswidth]{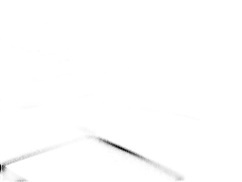}}}~
\subfloat[\label{fig:checkers_20}]{\frame{\includegraphics[width=\checkerswidth]{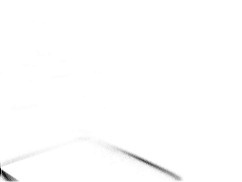}}}
    \caption{\label{fig:checkers} Sequence from a camera translating past a checkerboard (\ref{fig:checkers_02}-\ref{fig:checkers_05}). These grayscale frames, provided by the DAVIS~\cite{Brandli14ssc} are not used by our method; they are just for visualization purposes.
    Each image in \ref{fig:checkers_06}-\ref{fig:checkers_20} shows the IWE of each cluster (15 clusters, optical flow motion models). \ref{fig:checkers_all} shows the segmented output (combined IWEs) in the accustomed colored format.}
\end{figure*}

\subsection{Continuum Depth Variation with High-Resolution Event-based Camera}
Due to the recent development of new, high resolution event-based cameras~\cite{Son17isscc}, we show the results of our method on the output of a Samsung DVS Gen3 sensor, with a spatial resolution of $640\times 480$ pixels.
In this experiment we show the segmentation of several scenes (a textured carpet, some leaves and a temple poster) as the camera moves.
Due to ego-motion induced parallax, there is a continuous gradient in the motion in the scene, i.e., the scenes present a continuum of depths.
As can be seen in Fig. \ref{fig:high-res-all}, our method works the same on high-resolution data.
\global\long\def\layerwidth{0.25\linewidth}
\begin{figure*}[t]
	\centering
	\begin{adjustbox}{max width=\textwidth}
	\bgroup
    \def\arraystretch{1}
    \setlength{\tabcolsep}{2pt}
    {\small
	\begin{tabular}{
	>{\centering\arraybackslash}m{0.5cm}
	>{\centering\arraybackslash}m{\layerwidth} 
	>{\centering\arraybackslash}m{\layerwidth}
	>{\centering\arraybackslash}m{\layerwidth}}
	 & Leaves & Carpet & Temple \\
		\rotatebox{90}{Scene}
		&
		\frame{\includegraphics[width=\linewidth]{images/guillermo_leaf/index.jpg}}
		&
		\frame{\includegraphics[width=\linewidth]{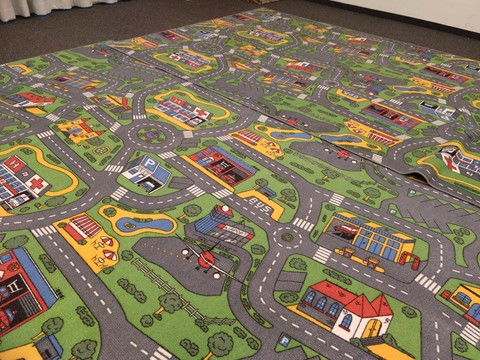}}
		&
		\frame{\includegraphics[width=\linewidth]{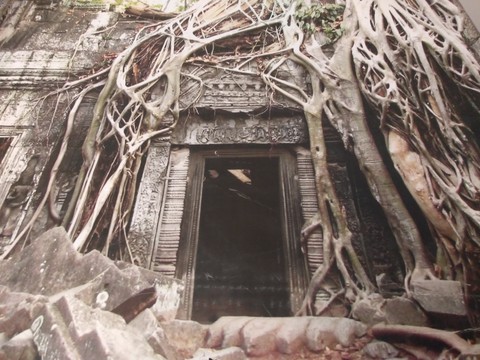}}
		\\
		\rotatebox{90}{Cluster 1}
		&
		\frame{\includegraphics[width=\linewidth]{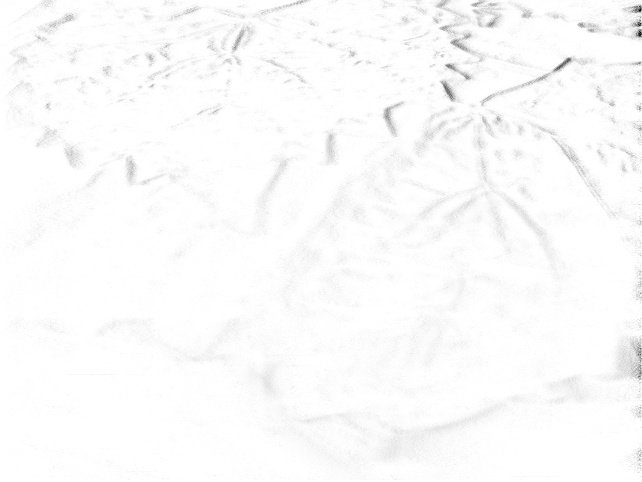}}
		&
		\frame{\includegraphics[width=\linewidth]{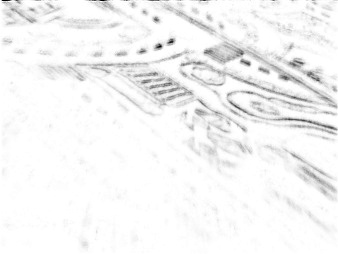}}
		&
		\frame{\includegraphics[width=\linewidth]{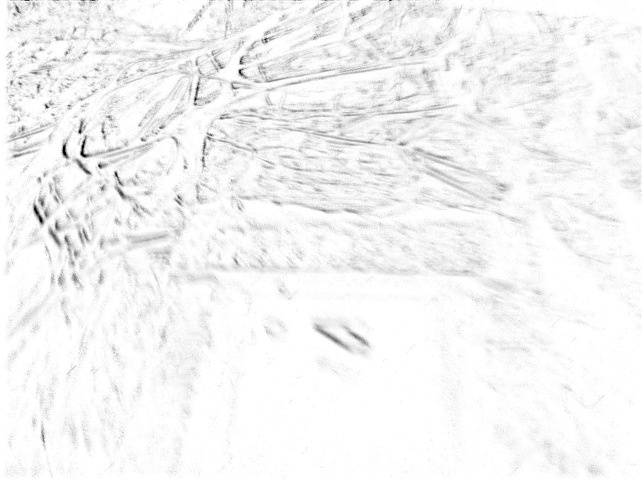}}
		\\	
		\rotatebox{90}{Cluster 2}
		&
		\frame{\includegraphics[width=\linewidth]{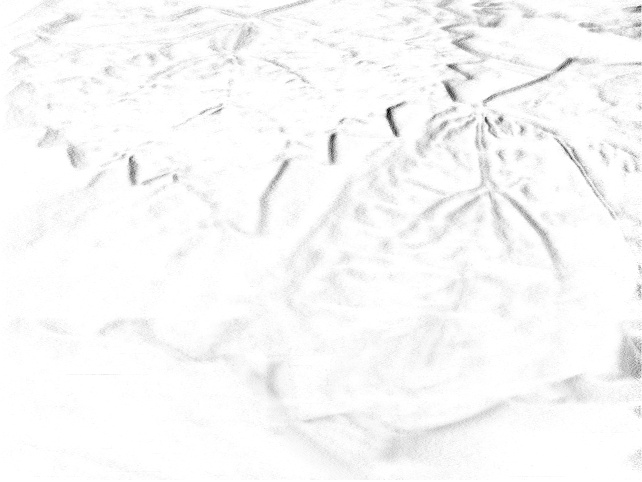}}
		&
		\frame{\includegraphics[width=\linewidth]{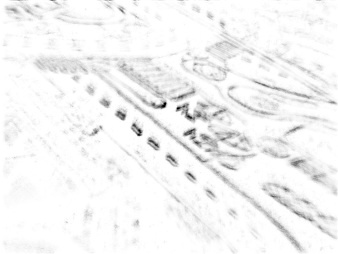}}
		&
		\frame{\includegraphics[width=\linewidth]{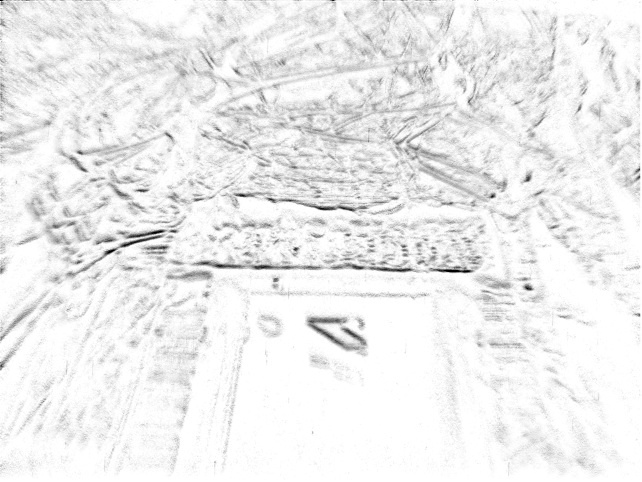}}
		\\
		\rotatebox{90}{Cluster 3}
		&
		\frame{\includegraphics[width=\linewidth]{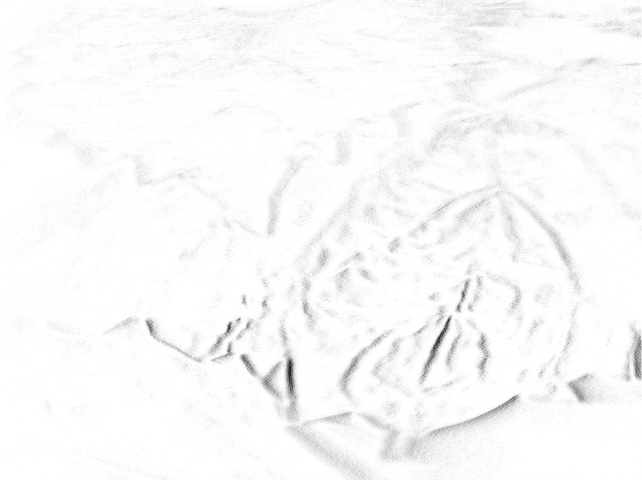}}
		&
		\frame{\includegraphics[width=\linewidth]{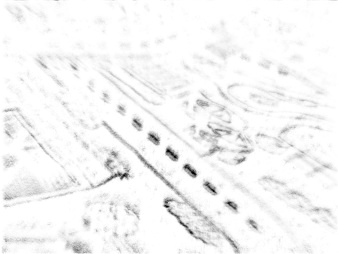}}
		&
		\frame{\includegraphics[width=\linewidth]{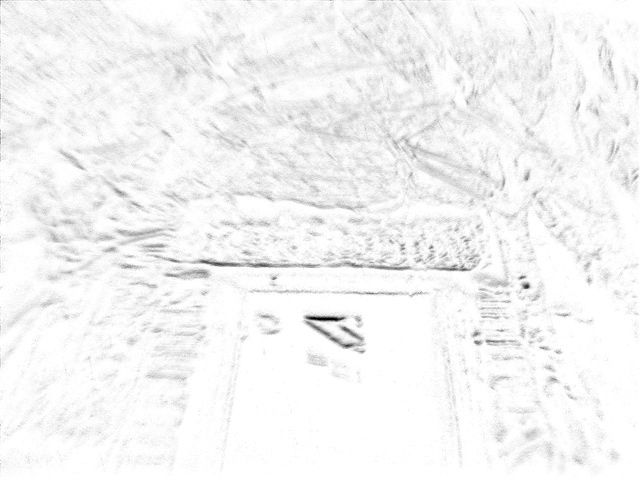}}
		\\
		\rotatebox{90}{$\ldots$}
		&
		\frame{\includegraphics[width=\linewidth]{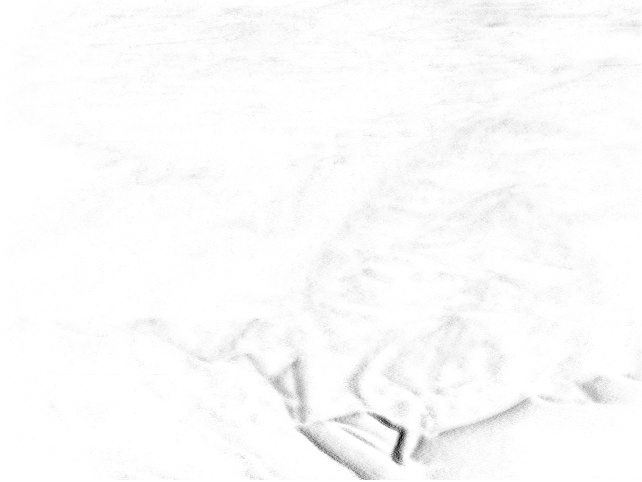}}
		&
		\frame{\includegraphics[width=\linewidth]{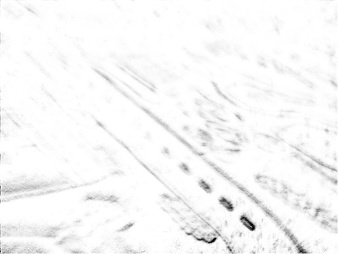}}
		&
		\frame{\includegraphics[width=\linewidth]{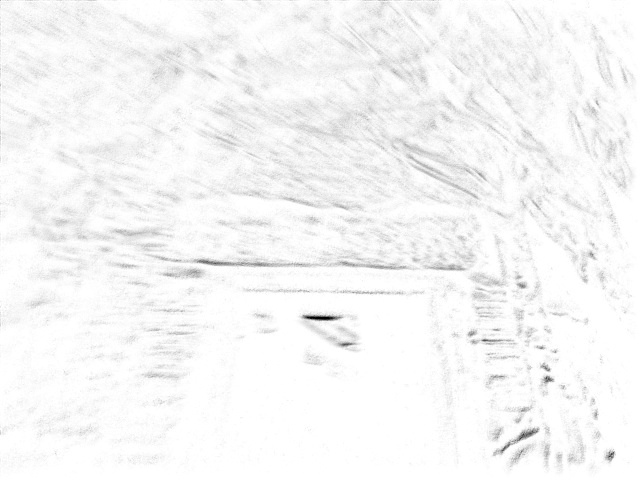}}
		\\
		\rotatebox{90}{Segmentation}
		&
		\frame{\includegraphics[width=\linewidth]{images/guillermo_leaf/output.jpg}}
		&
		\frame{\includegraphics[width=\linewidth]{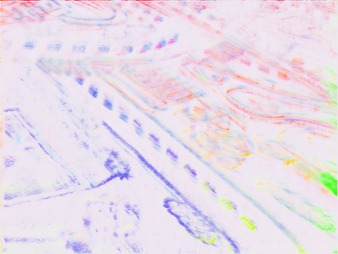}}
		&
		\frame{\includegraphics[width=\linewidth]{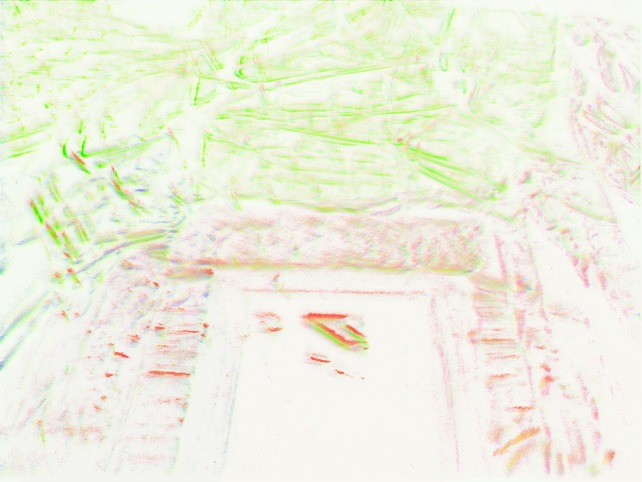}}
		\\
	\end{tabular}
	}
	\egroup
	\end{adjustbox}
	\caption{Scenes recorded with a Samsung DVS Gen3 event camera ($640 \times 480$ pixels); algorithm run with ten clusters ($\numLayers=10$). 
	From top to bottom: scene recorded, 
	four of the clusters (motion-compensated IWEs, with darkness indicating event likelihoods), 
	and the accustomed colored segmentation (as in Fig.~\ref{fig:method:layers}).
	These examples illustrate that our method can be used to segment the scene according to depth from the camera, although it is not its main purpose.}
	\label{fig:high-res-all}
\end{figure*}

\subsection{Comparison to k-means Optic Flow Clustering}
\label{subsec:exp:linear_slider}
Finally, the following experiments shows the comparison of our method against \mbox{k-means} clustering of optic flow.
We first illustrate the difference with a qualitative example and then quantitatively show the ability of our method to resolve small differences in velocities compared to \mbox{k-means}.
To this end, we use an event-based camera mounted on a motorized linear slider, which provides accurate ground truth position of the camera. 
Since the camera moves at constant speed in a 1-D trajectory, the differences in optical flow values observed when viewing a static scene are due to parallax from the different depths of the objects causing the events.

\paragraph{Numbers Sequence.}
\label{subsubsec:exp:ls:numbers}
In this experiment, we placed six printed numbers at different, known depths with respect to the linear slider.
The event-based camera moved back and forth on the slider at approximately constant speed.
Due to parallax, the objects at different depths appear to be moving at different speeds; faster the closer the object is to the camera.
Thus we expect the scene to be segmented into six clusters, each corresponding to a different apparent velocity.

\begin{figure}[t]
    \centering
    \subfloat[DAVIS performs linear translation over a multi-object scene.]{\includegraphics[width=0.47\linewidth]{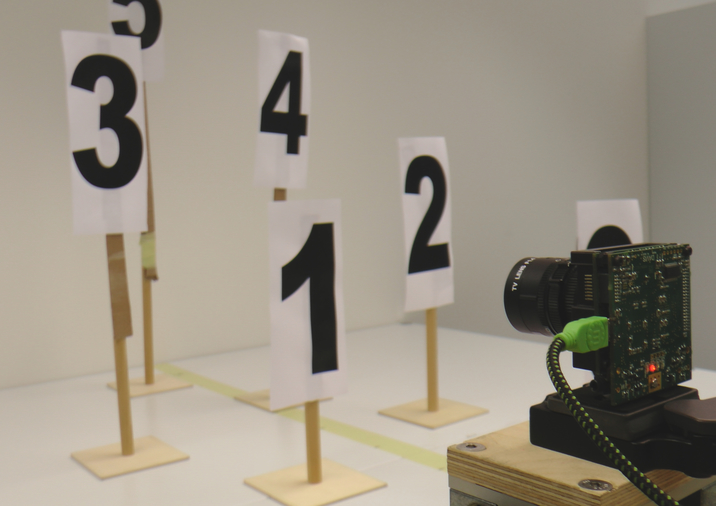}}~
    \subfloat[Resulting image and events (red and blue, indicating polarity).]{\includegraphics[width=0.47\linewidth]{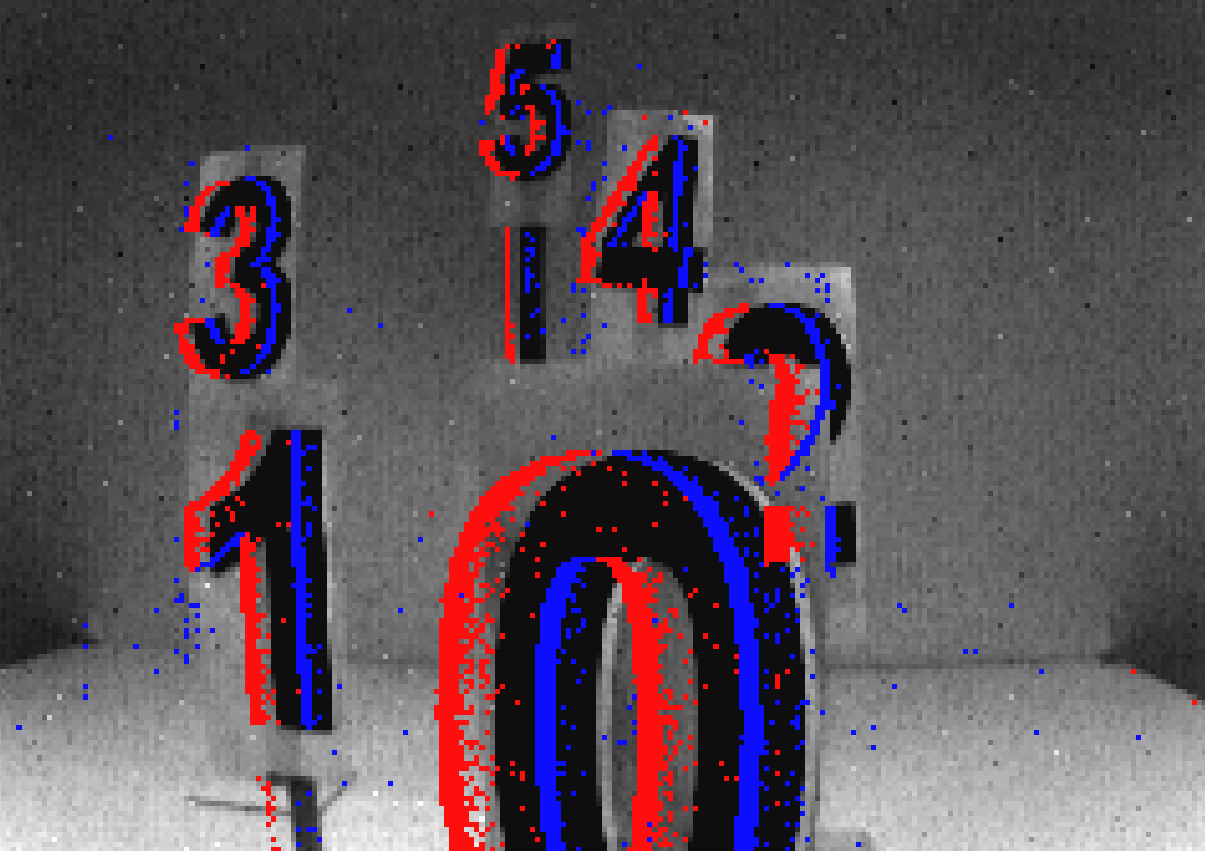}}\\[-0.5ex]
    \subfloat[Clustered volume of events (colored by cluster number).]{\includegraphics[width=0.47\linewidth]{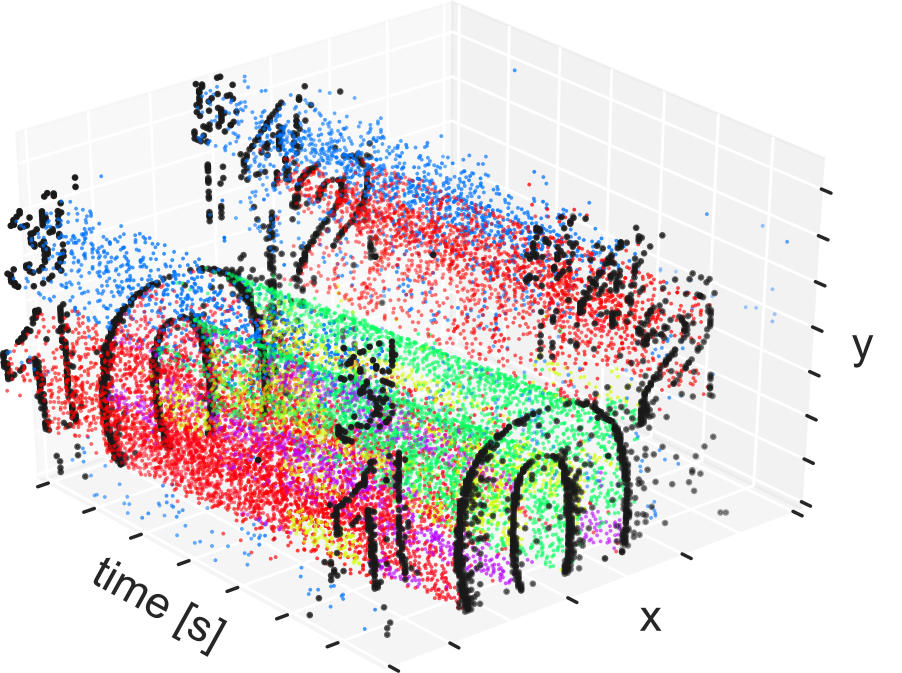}}~
     \subfloat[Motion-compensated image (colored by clustered optic flow).\label{fig:numbers:layers:IWEbad}]{\includegraphics[width=0.47\linewidth]{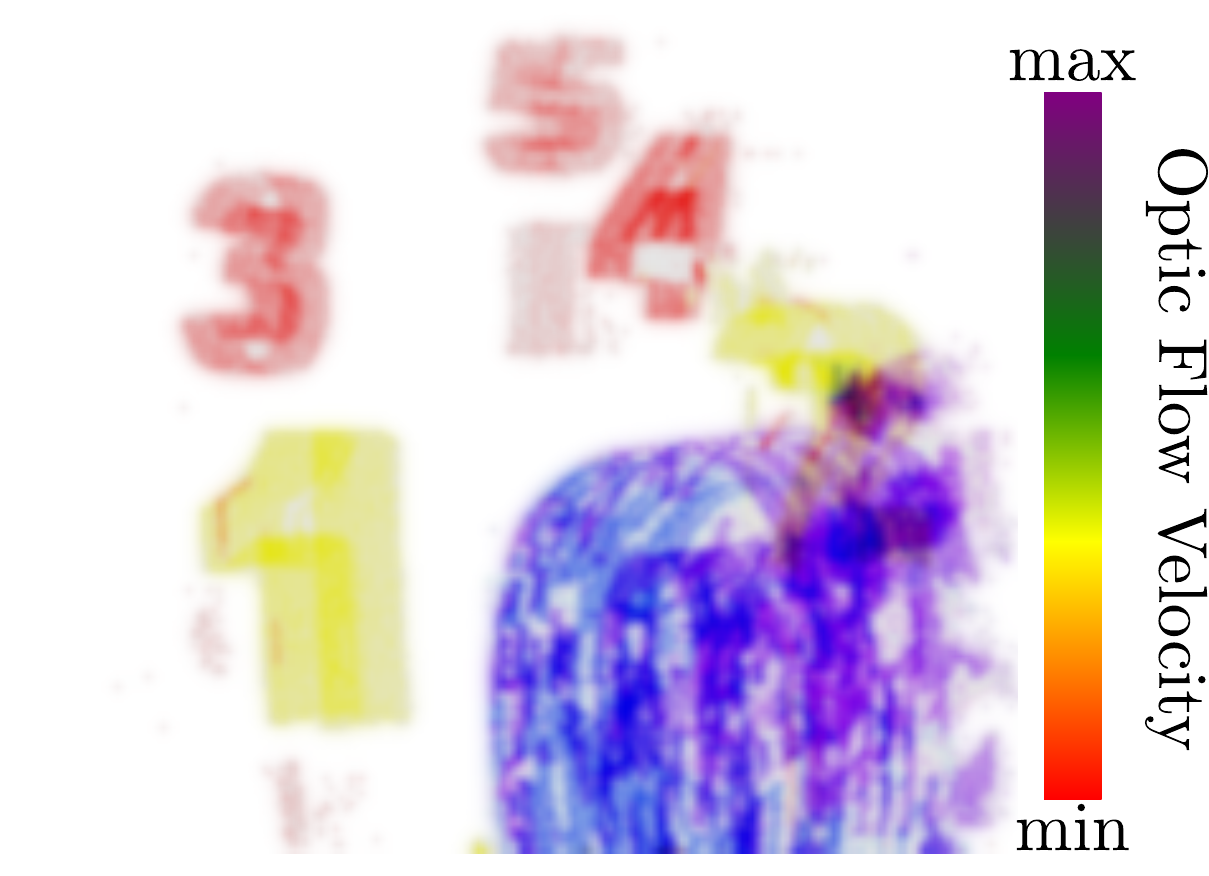}}\\[-0.5ex]     
    \subfloat[Clustered volume of events (colored by cluster number).]{\includegraphics[width=0.47\linewidth]{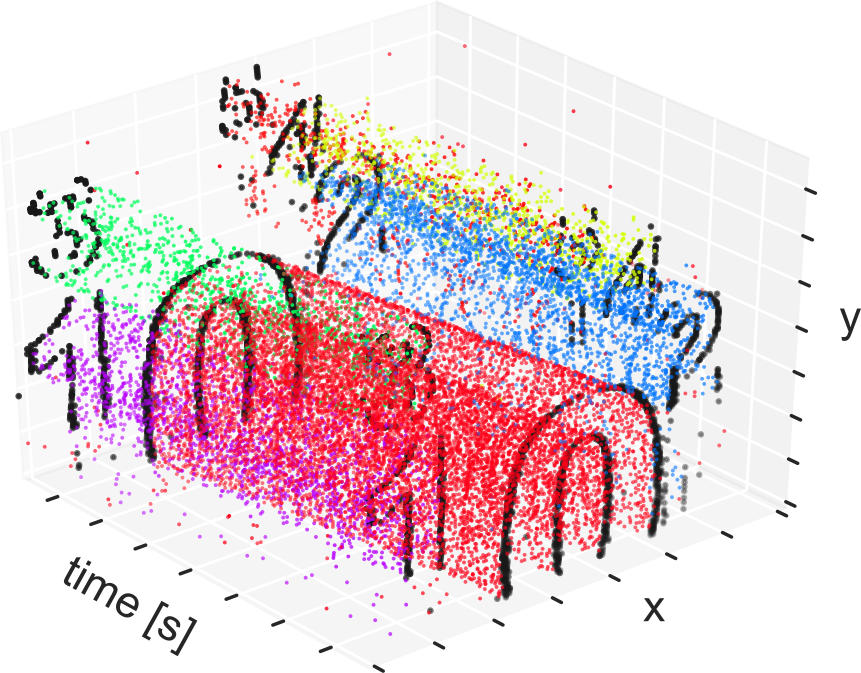}}~
    \subfloat[Motion-compensated image (colored by recovered optic flow).\label{fig:numbers:layers:IWEgood}]{\includegraphics[width=0.47\linewidth]{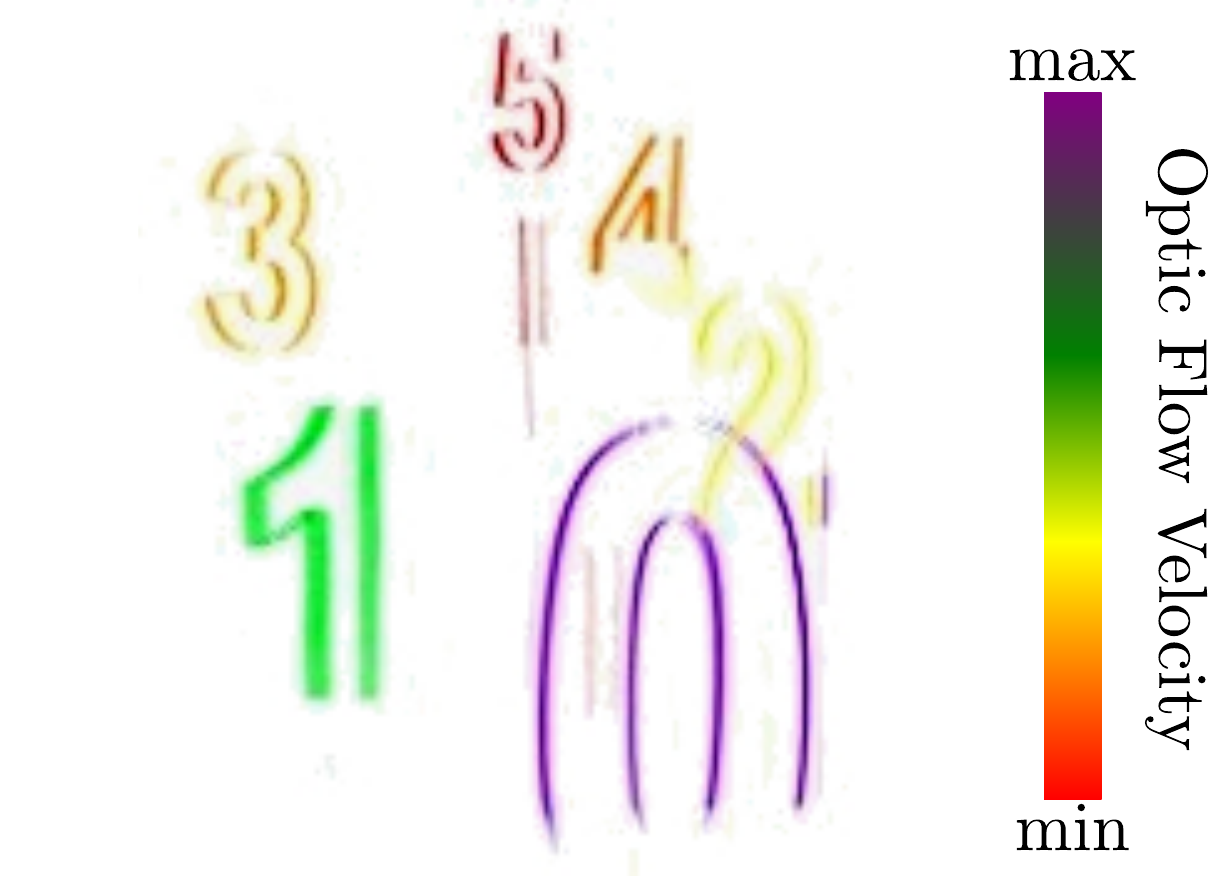}}
    \caption{\emph{Numbers Sequence}. 
    Motion Segmentation by \mbox{k-means} clustering on estimated optic flow (center row) 
    and by Algorithm~\ref{alg:ems} (bottom row).
    }
    \label{fig:numbers:layers}
\end{figure}

Fig.~\ref{fig:numbers:layers} compares the results of \mbox{k-means} clustering optic flow and Algorithm~\ref{alg:ems}.
To compute optical flow we use conventional methods on reconstructed images at a high frame rate~\cite{Scheerlinck18accv}, 
with the optical flow method in~\cite{Farnebaeck03scia} producing better results on such event-reconstructed images than state-of-the-art learning methods~\cite{Sun18cvpr}.
The results show that the velocities corresponding to the six numbers are too similar to be resolved correctly by the two-step approach (flow plus clustering), as evidenced by the bad segmentation of the scene (numbers 3, 4 and 5 are clustered together, whereas three clusters are used to represent the events of the fastest moving number--the zero, closest to the camera).
In contrast, our method accurately clusters the events according to the motion of the objects causing them, in this case, according to velocities, since we used an optical flow warp (linear motion on the image plane).
The higher accuracy of our method is easily seen in the sharpness of the motion-compensated images (cf. Fig.~\ref{fig:numbers:layers:IWEbad} and Fig.~\ref{fig:numbers:layers:IWEgood}).
%

\begin{figure}[t]
\centering
    \subfloat[\label{fig:rocks6:flowclusters:vornoi50}Minimum velocity between objects: \SI{50}{pixel/\second}]{\includegraphics[width=0.75\linewidth]{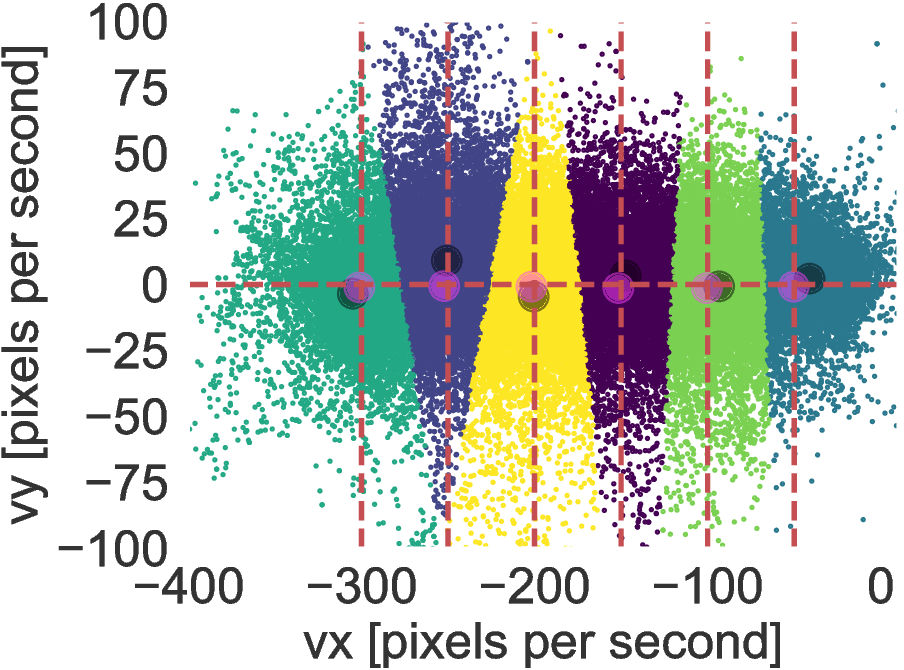}}\\
    \subfloat[\label{fig:rocks6:flowclusters:vornoi6}Minimum velocity between objects: \SI{6}{pixel/\second}]{\includegraphics[width=0.75\linewidth]{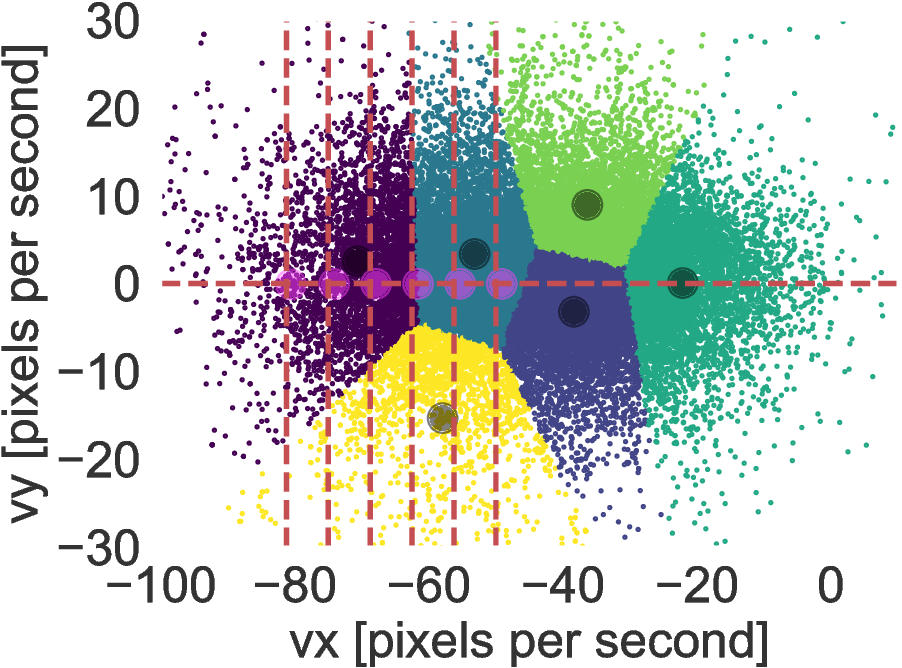}}
    \caption{\emph{Rocks at Different Speeds}. 
    Segmentation by \mbox{k-means} clustering of estimated optical flow ($k=6$).
    The plots show the distribution of optical flow vectors and the six Voronoi diagrams resulting from \mbox{k-means} clustering on optic flow space.
    The crossings of red dashed lines indicate the ground truth optical flow velocity, the dark circles indicate the centroids of the \mbox{k-means} clusters. 
    The pink circles indicate the cluster's optical flow estimated by our method (Algorithm~\ref{alg:ems}).}
    \label{fig:rocks6:flowclusters}
\end{figure}

\paragraph{Rocks at Different Speeds.}
\label{subsubsec:exp:ls:rocks}
We also tested our algorithm on two real sequences with six objects of textured images of pebbles (qualitatively similar to Fig.~\ref{fig:metric}), 
in which the relative velocities of the objects were \SI{50}{pixels/\second} and \SI{6}{pixels/\second}, respectively.
Fig.~\ref{fig:rocks6:flowclusters} shows the results. 
If the objects are moving with sufficiently distinct velocities (Fig.~\ref{fig:rocks6:flowclusters:vornoi50}), the clusters can be resolved by the two-step approach. 
However, once the objects move with similar velocities (Fig.~\ref{fig:rocks6:flowclusters:vornoi50}), \mbox{k-means} clustering of optical flow is unable to correctly resolve the different clusters.
In contrast, our method works well in both cases; it is much more accurate: 
it can resolve differences of \SI{6}{pixel/\second} for objects  moving at \SIrange{50}{80}{pixel/\second}, given the same slice of events.

\cleardoublepage
\balance
\small 
{
	\bibliographystyle{ieeetr_fullname} 

}
\end{document}